\pdfoutput=1
\documentclass[sigconf]{acmart}
\usepackage{algorithm}
\usepackage{algorithmic}
\usepackage{amsmath,amssymb}

\usepackage{multirow}
\usepackage{subcaption}
\DeclareMathOperator*{\argmax}{arg\,max}

\def\BibTeX{{\rm B\kern-.05em{\sc i\kern-.025em b}\kern-.08emT\kern-.1667em\lower.7ex\hbox{E}\kern-.125emX}}

\begin{document}

	%
	\title{Attentive One-Dimensional Heatmap Regression for Facial Landmark Detection and Tracking} 

	%

\title{Attentive One-Dimensional Heatmap Regression for Facial Landmark Detection and Tracking} 
\author{ Shi Yin$^1$, Shangfei Wang$^\ast$$^{1,2}$, Xiaoping Chen$^2$, Enhong Chen$^3$ and Cong Liang$^1$}
\email{davidyin@mail.ustc.edu.cn, {sfwang, xpchen, cheneh}@ustc.edu.cn, lc150303@mail.ustc.edu.cn}
\affiliation{
	\institution{$^1$Key Lab of Computing and Communication Software of Anhui Province,\\School of Computer Science and Technology, University of Science and Technology of China}
	\institution{$^2$Anhui Robot Technology Standard Innovation Base, University of Science and Technology of China}
	\institution{$^3$Anhui Province Key Lab of Big Data Analysis and Application,\\School of Computer Science and Technology, University of Science and Technology of China}
	\institution{$^\ast$Dr. Shangfei Wang is the corresponding author.}
}
	
	%

	%
	
\begin{abstract}
	Although heatmap regression is considered a state-of-the-art method to locate facial landmarks, it suffers from huge spatial complexity and is prone to quantization error.  To address this, we propose a novel attentive one-dimensional heatmap regression method for facial landmark localization. First, we predict two groups of 1D heatmaps to represent the marginal distributions of the $x$ and $y$ coordinates. These 1D heatmaps reduce spatial complexity significantly compared to current heatmap regression methods, which use 2D heatmaps to represent the joint distributions of $x$ and $y$ coordinates. With much lower spatial complexity, the proposed method can output high-resolution 1D heatmaps despite limited GPU memory, significantly alleviating the quantization error. Second, a co-attention mechanism is adopted to model the inherent spatial patterns existing in $x$ and $y$ coordinates, and therefore the joint distributions on the $x$ and $y$ axes are also captured. Third, based on the 1D heatmap structures, we propose a facial landmark detector capturing spatial patterns for landmark detection on an image; and a tracker further capturing temporal patterns with a temporal refinement mechanism for landmark tracking. Experimental results on four benchmark databases demonstrate the superiority of our method.  
\end{abstract}



\maketitle

\section{Introduction}
\begin{figure}
	\centering
	\includegraphics[scale=0.44]{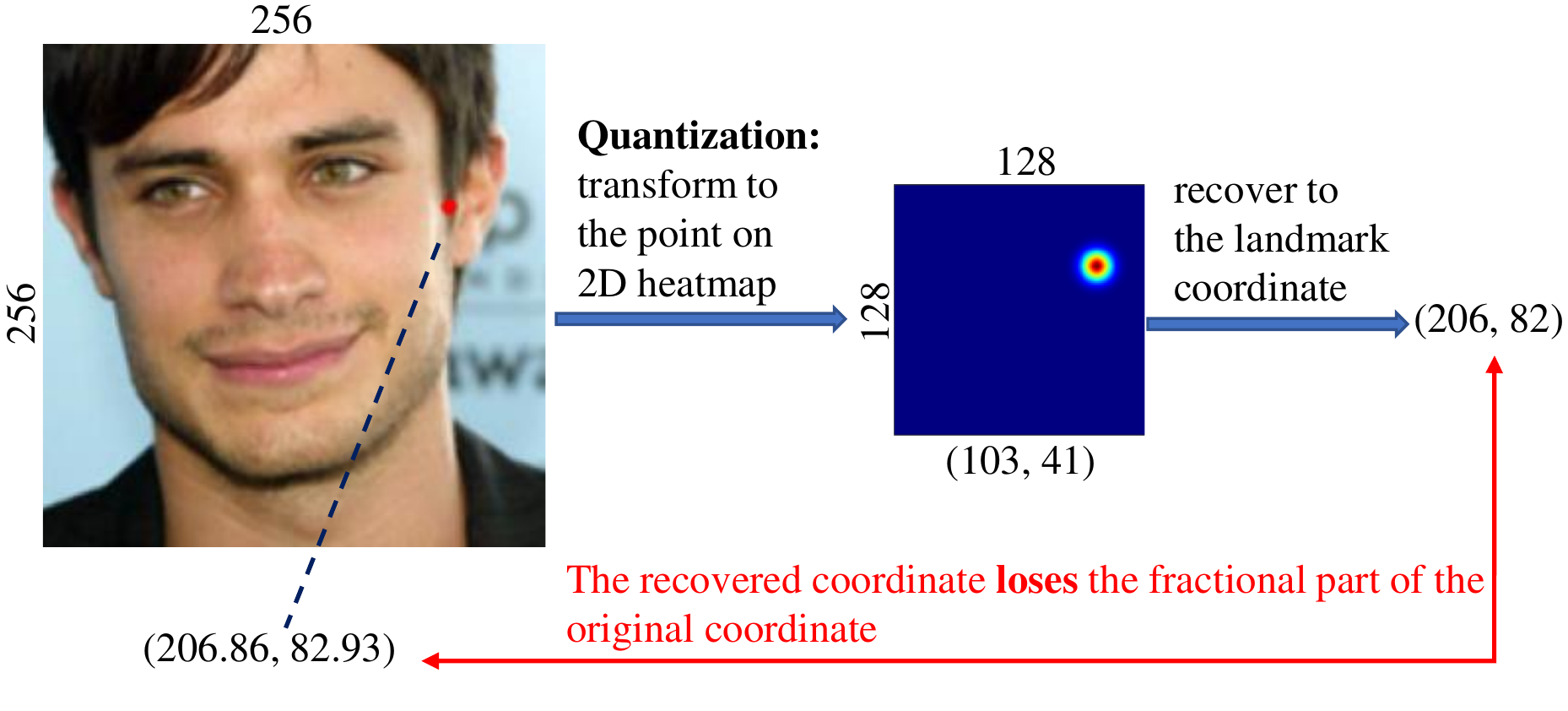}
	\caption{Illustration of the quantization error caused by low-resolution 2D heatmap. The facial image is sampled from the 300W \cite{sagonas2016300} dataset.}
	\label{imgimg1}
\end{figure}

As a fundamental computer vision task, face alignment \cite{survey,survey2} consists of two sub-tasks, i.e., facial landmark detection on a static image, and facial landmark tracking in a video with continuous frames. Regression-based methods for face alignment can be divided into two categories, i.e., a coordinate regression approach, or a heatmap regression approach. Coordinate regression approaches directly map facial appearances to the continuous values of landmark coordinates or their displacements. This technique has difficulty handling spatial distributions around ground truth coordinates, especially for the coordinates with large variances. Therefore, as reported by Sun ~\emph{et al.} \cite{better_heatmap1}, it is sub-optimal on spatial generalization performance.

Recently, heatmap regression approaches have been proposed.  These approaches assume that a landmark coordinate obeys Gaussian distribution around its ground truth label. These approaches predict the discrete probability distribution that a landmark occurs in each point of the heatmap, and transform points on the heatmaps to the predicted landmark coordinates. Heatmap regression approaches successfully capture the spatial distributions around groun-d truths. They are thus theoretically capable of better spatial modeling than coordinate regression approaches. However, the quantization process, i.e., using discrete heatmaps to represent continuous landmark coordinates with both integer and fractional parts, may cause quantization error. One way to alleviate such a problem is to improve the heatmap resolution. Unfortunately, current heatmap regression methods suffer from high spatial complexity, as they use 2D heatmaps to represent joint distributions on the $x$ and $y$ axes. Specifically, a landmark produces a heatmap with $L \times L$ points, where $L$ is the output resolution on the $x$ and $y$ axes. The total output size of $N$ landmarks is $NL^2$. The space occupation of 2D heatmaps increases dramatically as $L$ increases. Due to the limited machine memory, $L$ is typically restricted to a value lower than the input face \cite{hourglass,how_far}. This  compromise may cause severe quantization errors as shown in Fig. \ref{imgimg1}, limiting the practical performance of the heatmap regression method. Although some methods \cite{better_heatmap1,tai2018towards,better_heatmap2,better_heatmap3} are proposed to estimate landmark coordinates with fractional parts from a 2D heatmap, they still suffer from the information loss caused by low resolution.

\begin{figure*}[t]
	\centering
	\begin{minipage}[t]{\linewidth}
		\centering
		\includegraphics[scale=0.498]{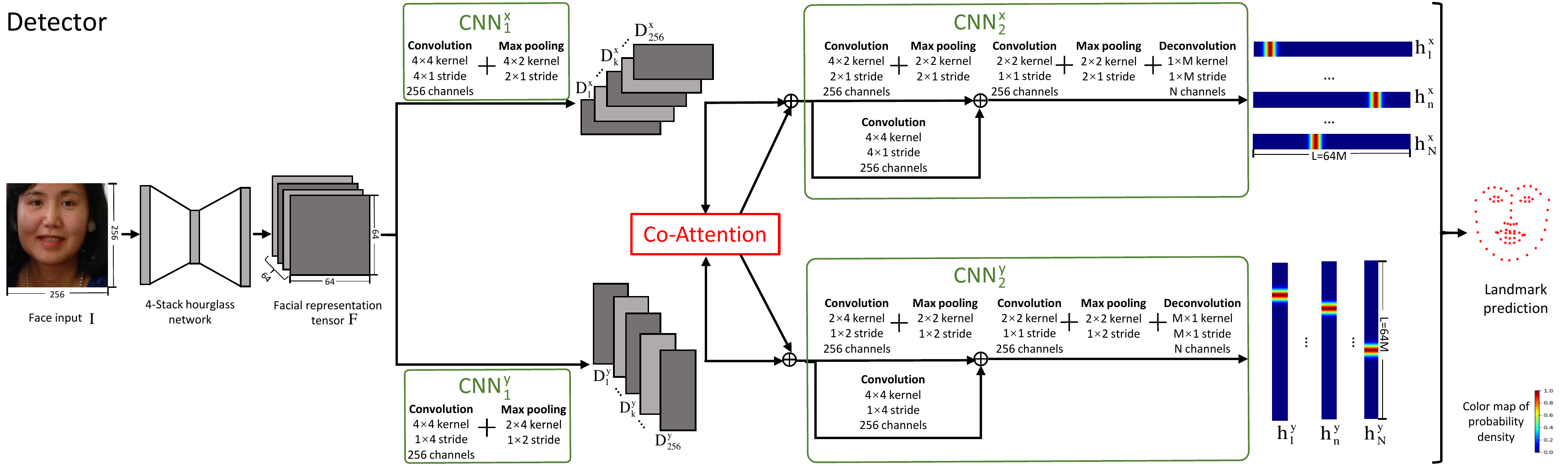}
		\subcaption{1D heatmaps on the $x$ and $y$ axes are generated by CNN networks. The co-attention mechanism is applied between the output features of $CNN_1^x$ and $CNN_1^y$ to capture joint distribution on the two axes. }
		\label{detector}
	\end{minipage}
	\begin{minipage}[t]{\linewidth}
		\centering
		\includegraphics[scale=0.499]{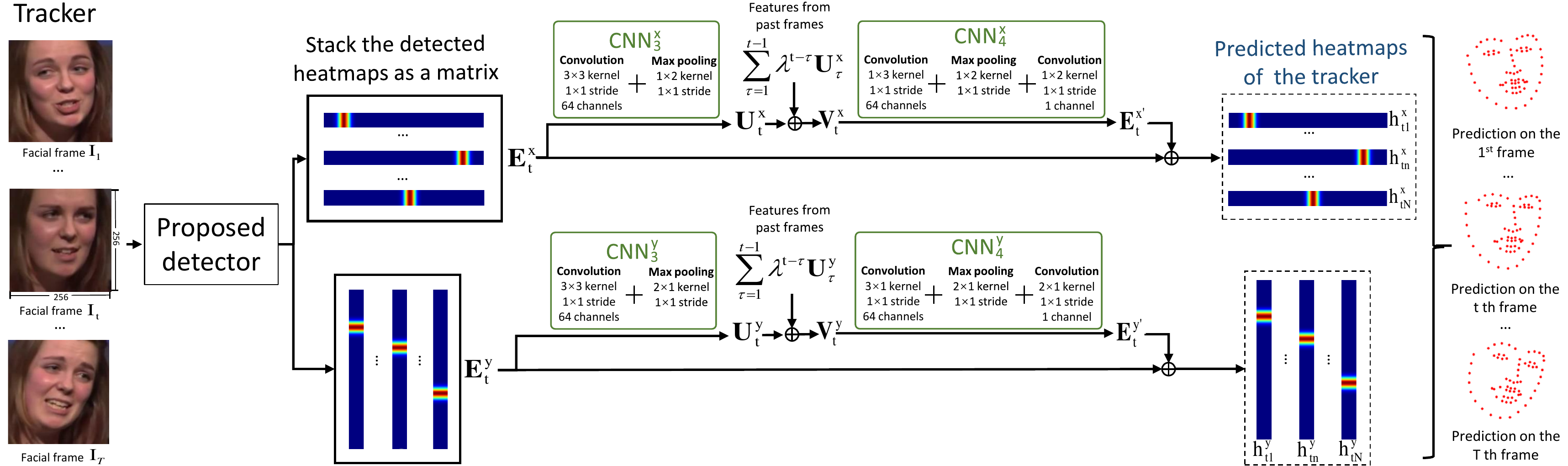}
		\subcaption{Based on the spatial patterns captured by the detector, the tracker further captures temporal patterns by fusing features on the current frame with features from past frames. Temporal patterns are used to refine the detected heatmaps.}
		\label{tracker}
	\end{minipage}
	\caption{The proposed detector (a) and tracker (b).}
\end{figure*}

To solve the disadvantage of 2D heatmaps, we propose a novel attentive one-dimensional heatmap regression  approach that achieves prominent spatial and temporal modeling capability while significantly decreasing the output complexity. The basic idea is to replace the 2D heatmaps by 1D heatmaps that represent marginal distributions on the $x$ and $y$ axes as the output structure. Considering the correlation between $x$ and $y$ coordinates, we capture the $x$-$y$ joint distribution implicitly by applying a co-attention mechanism on the distributional features of the two axes. Our method significantly decreases the spatial complexity of the output. The total output size of 1D heatmaps on two axes is $2NL$ ($N\times2L$), much smaller than the 2D heatmaps ($NL^2$). The small output size allows us to fully boost the resolution despite limited GPU memory, significantly alleviating quantization errors. 
Based on the proposed heatmap structure, we design a facial landmark detector and a tracker, as shown in Fig.  \ref{detector} and \ref{tracker}, respectively. The detector captures spatial patterns in a static image, while the tracker integrates both spatial and temporal patterns from the facial sequence in a video to locate landmarks. For the detector, we design two groups of convolutional neural networks (CNNs) to compress the facial representation tensor to 1D heatmaps on the $x$ and $y$ axes. The co-attention mechanism is adopted between the features extracted by the CNNs. In the tracker, spatial patterns from the current frame are extracted by the proposed detector, temporal patterns among multiple frames are embedded by a novel temporal refinement mechanism integrating features on the current frame and features from past frames.

We conduct facial landmark detection experiments on the 300W dataset and the AFLW dataset, and conduct facial landmark tracking experiments on the 300VW dataset and the TF dataset.  Experimental results on these datasets show that the proposed method outperforms state-of-the-art coordinate regression and heatmap regression methods.

The main contributions of our method are three folds. First, we are the first that propose to predict 1D heatmaps on the $x$ and $y$ axes instead of using 2D heatmaps to locate landmarks and successfully alleviate the quantization error with a fully boosted output resolution. Second, we propose a co-attention module to capture the joint coordinate distribution on the two axes.  Third, based on the proposed heatmap regression method, we design a facial landmark detector and tracker which achieve state-of-the-art performance.

\section{Related Work}
\subsection{Coordinate Regression Methods}
Coordinate regression approaches \cite{SDM,ESR,Xavier,Ren,Cascade_CNN,TCDCN,TSTN,RBM} directly predict coordinate values or their increments by a mapping function. As early works, Xiong ~\emph{et al.} \cite{SDM} proposed a Supervised Descent Method (SDM) method, which maps Scale-Invariant Feature Transform (SIFT) features to landmark displacements between the current output and the ground truth by minimizing a nonlinear least square objective function.  Cao ~\emph{et al.} \cite{ESR} and Burgos-Artizzu ~\emph{et al.} \cite{Xavier} learned fern regressors to predict landmark increments. Ren ~\emph{et al.} \cite{Ren} proposed to learn local binary features for every facial landmark by random forests. 

Recently, deep learning based methods are proposed for face alignment. Sun ~\emph{et al.} \cite{Cascade_CNN} proposed a CNN based method to predict facial landmarks in a cascaded way. Zhang ~\emph{et al.} \cite{TCDCN} proposed Tasks-Constrained Deep Convolutional Network (TCDCN), a multi-task learning method that learns to predict landmark coordinates as well as other facial attributes, including expression, gender, etc.
Liu ~\emph{et al.} \cite{TSTN} combined a CNN network with a RNN based encoder-decoder network to learn spatial and temporal patterns of landmarks in adjacent frames. 

To capture dependencies among landmark labels, some works adopted probabilistic graphical models, such as a Dynamic Bayesian Network (DBN) \cite{DBN}, or a Restricted Boltzmann Machine (RBM) \cite{RBM}, as shape constraints. Yin ~\emph{et al.} \cite{GAN_Tracking} utilized adversarial learning to explore the inherent dependencies among the movement of facial landmarks.

Despite these progresses, it is still challenging for a coordinate regression method to capture spatial distributions around ground truth coordinates, especially when the coordinate values vary in a wide range. This weakness leads to a sub-optimal spatial generalization performance of these methods.

\subsection{Heatmap Regression Methods}
Heatmap regression methods \cite{hourglass,how_far,encoder_decoder,Chu,LAB,CNNCRF,better_heatmap1,tai2018towards} capture spatial distributions around ground truths by likelihood heatmaps of landmarks. Newell ~\emph{et al.} \cite{hourglass} proposed a stacked hourglass network to generate heatmaps for 2D human pose estimation. Bulat and Tzimiropoulos \cite{how_far} enhanced 
the stacked hourglass network with hierarchical, parallel and multi-scale residual blocks. Xi ~\emph{et al.} \cite{encoder_decoder} proposed a spatial and temporal recurrent learning method for landmark detection and tracking. Based on the heatmap technique, Chu ~\emph{et al.} \cite{Chu} proposed a multi-context attention mechanism to focus on informative feature regions. Wu ~\emph{et al.} \cite{LAB} proposed to estimate the heatmap of facial boundary as auxiliary features to locate landmarks. Chen ~\emph{et al.} \cite{chen2019adversarial} designed an adversarial learning method to learn structural patterns among landmarks. Chen ~\emph{et al.} \cite{CNNCRF} proposed a Conditional Random Field (CRF) method to embed geometric relationships among landmarks based on their heatmaps. Liu ~\emph{et al.} \cite{Liu_2019_CVPR} proposed a heatmap correction unit which uses global shape constraints to refine heatmaps.

The heatmap structure achieves good theoretical performance for spatial generalization. However, it suffers from huge spatial complexity. Under limited space, the heatmap resolution is typically compressed to a value smaller than the input face. Consequently, fractional parts of landmark coordinates are totally dropped, leading to serious quantization errors. To estimate the landmark coordinates with fractional parts based on 2D discrete heatmaps, Sun ~\emph{et al.} \cite{better_heatmap1} proposed to integrate all point locations weighted by their probabilities in the 2D heatmap as the predicted coordinate. Nibali ~\emph{et al.} \cite{better_heatmap2} proposed a differentiable layer at the end of CNN networks to convert heatmaps as continuous coordinates. Tai ~\emph{et al.} \cite{tai2018towards} proposed Fractional Heatmap Regression (FHR), which uses three heatmap points to predict the fractional parts of landmark coordinates according to a Gaussian function. Zhang ~\emph{et al.} \cite{better_heatmap3} proposed to estimate the landmark coordinates by a Taylor-expansion based distribution approximation. However, for these 2D heatmap-based methods, when the heatmap resolution is low, the heatmap points are sparse and some detailed information on the spatial distribution is lost, making it very hard for coordinate estimation.

To address the disadvantages of 2D heatmap regression, we propose a new regression method based on 1D heatmaps which represent the marginal distribution on each axis. We capture joint distributions between the $x$ and $y$  axes by a co-attention mechanism, instead of using 2D heatmaps. The proposed regression method is much more space-efficient and the output resolution can be fully boosted despite limited space. Therefore, the quantization error is significantly alleviated.

\section{Analysis on the Quantization Error of 2D Heatmap} 

Conventional heatmap regression methods, denoted as $f(\cdot)$, predict discrete 2D joint distributions (heatmaps) for $N$ predefined landmarks from a facial image $\textbf{I} \in\mathbb{R}^{F \times F \times 3}$, as shown in Equation \eqref{ps}:
\begin{equation}
\label{ps}
\textbf{H}_n = f(\textbf{I},n;\theta_{f}), 1 \leq n \leq N
\end{equation}
where $\theta_{f}$ is the parameters of $f(\cdot)$ and $\textbf{H}_{n}\in\mathbb{R}^{L \times L}$ denotes the heatmap of the $n$ $ (1\leq n\leq N)$ th landmark with resolution $L$. The ground truth heatmap, represented as $\textbf{H}^*_{n}$, is considered a discrete Gaussian distribution centered on the ground truth position, i.e., $(x^*_n, y^*_n)$. The probability density on an arbitrary heatmap point $(x, y)$ follows Equation \eqref{hm}:
\begin{equation}
\label{hm}
\textbf{H}^*_{n}(x,y) \varpropto  exp(-\frac{1}{2\sigma^2}((x-x^*_n)^2+(y-y^*_n)^2)) 
\end{equation}
where $\sigma$ is the variance of the Gaussian distribution.

The heatmap structure models the spatial distributions of landmarks to obtain spatial generalization. However, using a discrete heatmap to represent continuous landmark coordinates with both integer and fractional parts may cause quantization error. This is because the rounding-down operation $\lfloor\,\rfloor$ is applied to convert the continuous coordinate $(p,q)$ to the discrete heatmap point $(x,y)$, as shown in Equation \eqref{round}:
\begin{equation}
\label{round}
x=\lfloor \frac{p \cdot L}{F} \rfloor , \,\,\, y=\lfloor \frac{q \cdot L}{F} \rfloor
\end{equation}
The rounding-down operation drops the fractional part of its input. Therefore, we could only recover an approximate value of $(p, q)$ from $(x,y)$, as shown in Equation \eqref{recover}:
\begin{equation}
\label{recover}
p'=x \cdot \frac{F}{L} =\frac{\lfloor  p \cdot (\frac{L}{F})\rfloor}{(\frac{L}{F})} ,\,\,\,\,\,\, q'=y \cdot \frac{F}{L} = \frac{\lfloor q \cdot (\frac{L}{F})\rfloor}{(\frac{L}{F})}\\
\end{equation}
where $(p', q')$ is the coordinate recovered from the heatmap. Quantization error $E$ is defined as the Euclidean distance between $(p, q)$ and $(p', q')$, as shown in Equation \eqref{quantization}:
\begin{equation}
\begin{split}
\label{quantization}
E=\sqrt{(p-p')^2+(q-q')^2}
\end{split}
\end{equation}
From Equation \eqref{round} and Equation \eqref{recover}, we find that the higher the value of $\frac{L}{F}$, the closer $p$ comes to $p'$ and $q$ comes to $q'$. In other words, the quantization error $E$ is decreasing. For example, suppose $(p, q)$ is $(142.84, 188.72)$. When $\frac{L}{F}=0.5$, $E=1.11$. When $\frac{L}{F}$ increases to $3.0$, $E$ decreases to $0.18$. 
Therefore, with a given $F$, one way to reduce $E$ is to improve $L$.

Unfortunately, the 2D heatmap structure is very spatially complex. For $N$ landmarks, a total of $NL^2$ heatmap points are generated. Due to the limited machine memory, current heatmap regression methods usually set $L$ as a value smaller than $F$, resulting in severe quantization errors.

\section{Methodology}
The key to reducing quantization errors is to boost the output resolution $L$ despite limited machine memory. For that purpose, we propose a new method with good capability of spatial and temporal modeling while significantly decreasing output complexity compared to 2D heatmap regression methods. Instead of predicting joint distributions on the $x$ and $y$ axes explicitly by 2D heatmaps that occupy huge space, we propose to model joint distributions implicitly and just predict 1D heatmaps that represent marginal distributions.

Based on such an idea, we propose a new landmark detector, as depicted in Fig. \ref{detector}. The detector $g_{de}(\cdot)$ with parameters $\theta_{de}$ is formalized as 
Equation \eqref{hl}, where $\textbf{h}_{n}^x$ and $\textbf{h}_{n}^y$ are the predicted 1D heatmaps on the $x$ and $y$ axes, respectively,  for the $n$ th landmark. 
\begin{equation}
\begin{split}
\label{hl}
\textbf{h}_n^x = g_{de}(\textbf{I},n,x;\theta_{de}),\,\,\,\,\textbf{h}_n^y = g_{de}(\textbf{I},n,y;\theta_{de}),\,\,\,\, 1 \leq n \leq N \\
\end{split}
\end{equation}
Coordinate prediction for the $n$ th landmark ($\hat{p}_n$,$\hat{q}_n$) on a facial image is obtained from the maximum points of $\textbf{h}_{n}^x$ and $\textbf{h}_{n}^y$, as shown in Equation \eqref{extreme}:
\begin{equation}
\begin{split}
\label{extreme}
\hat{p}_n=\argmax (\textbf{h}_{n}^x)\cdot\frac{F}{L},\,\,\,\,\, \hat{q}_n=\argmax (\textbf{h}_{n}^y)\cdot\frac{F}{L}
\end{split}
\end{equation}

We also extend the detector as a tracker, as depicted in Fig. \ref{tracker}. The tracker $g_{tr}(\cdot)$ with parameters $\theta_{tr}$ predicts landmark positions in a facial video, denoted as $\textbf{I}_{1:T}=(\textbf{I}_{1},...,\textbf{I}_{t},...,\textbf{I}_{T})$,  where $\textbf{I}_{t}$ is the $t$ $(1\leq t\leq T)$ th frame of the video. For the $t$ th frame, the tracker captures not only spatial patterns on $\textbf{I}_{t}$, but also temporal patterns inherent in the sequence from $\textbf{I}_{1}$ to $\textbf{I}_{t}$. The tracker is formalized as Equation \eqref{hl_t}.
\begin{equation}
\begin{split}
\label{hl_t}
&\textbf{h}_{tn}^x = g_{tr}(\textbf{I}_{1:t}, n,x;\theta_{tr}) ,\\ 
&\textbf{h}_{tn}^y = g_{tr}(\textbf{I}_{1:t},n,y;\theta_{tr}), 1 \leq n \leq N,1 \leq t \leq T \\
\end{split}
\end{equation}
where $\textbf{h}_{tn}^x$ and $\textbf{h}_{tn}^y$ are the output heatmaps on the $t$ th frame.

The detector and the tracker output $\textbf{h}_{n}^x$ and $\textbf{h}_{n}^y$ (or $\textbf{h}_{tn}^x$ and $\textbf{h}_{tn}^y$) $\in\mathbb{R}^L$  for $N$ landmarks, and the total output size for a facial image or a facial frame is $N\times 2L =2NL$, much smaller than that of the 2D heatmaps ($NL^2$). In other words, the light-weight structure of 1D heatmap allows us to boost its resolution $L$ to a large value without heavy space occupation, and therefore the quantization error is significantly alleviated.

\subsection{Detector}
\subsubsection{Generating 1D Heatmaps Using CNNs}
First, a facial representation $\textbf{F}$ is learned from the input face through a stacked hourglass network \cite{how_far}, as shown in the left part of Fig. \ref{detector}. Then, 1D heatmaps on two axes are generated by two groups of CNNs, respectively, as depicted in the green border boxes of Fig. \ref{detector}. The first group of CNNs, composed of $CNN_1^x$ and $CNN_2^x$, converts $\textbf{F}$ to 1D heatmaps on the $x$ axis, i.e., $\textbf{h}^{x}_{1},...,\textbf{h}^{x}_{n},...,\textbf{h}^{x}_{N}$, by compressing features along the $y$ axis with a striding operation. At the end of $CNN_2^x$, a deconvolution module is adopted to generate heatmaps and the heatmap resolution is proportional to the kernel and stride size ($M$) of deconvolution. The second group of CNNs,  composed of $CNN_1^y$ and $CNN_2^y$, generates heatmaps on the $y$ axis by compressing features along the $x$ axis.

\subsubsection{Capturing Joint Distribution on the $x$ and $y$ Axes by Co-Attention}
Co-attention \cite{co-attention} is a category of attention methods which capture the correlation between pairwise features. As shown in the red border box of Fig. \ref{detector}, we design a co-attention module to capture the $x$-$y$ joint distributions of landmark coordinates, . First, we encode correlations between features representing distributions on the $x$ and $y$ axes as affinity matrices. Second, features are converted by the affinity matrices and then fused together to embed the joint distribution into their representations.

The co-attention mechanism is adopted between the output features of $CNN_1^x$ and $CNN_1^y$, denoted as $\textbf{D}^x$ and $\textbf{D}^y$, respectively. Both $\textbf{D}^x$ and $\textbf{D}^y$ have multiple channels, $\textbf{D}^x_k$ and $\textbf{D}^y_k$ represent their $k$ th channel.  $\textbf{D}^x_k$ is a feature representing the distribution on the $x$ axis, while $\textbf{D}^y_k$ represents the distribution on the $y$ axis. The shape of  $\textbf{D}^x_k$ is the same as $(\textbf{D}^y_k)^\mathrm{T}$.
Two affinity matrices, i.e., $\textbf{W}_k^{xy}$ and $\textbf{W}_k^{yx}$, are adopted to encode the correlation between $\textbf{D}^x_k$ and $\textbf{D}^y_k$, as shown in Equation \eqref{att1}:
\begin{equation}
\begin{split}
\label{att1}
&\textbf{W}_k^{xy}=softmax(\frac{(\textbf{D}^{y}_k)^\mathrm{T}\cdot\textbf{P}\cdot(\textbf{D}^{x}_k)^\mathrm{T}}{\sqrt{d}}),\\
&\textbf{W}_k^{yx}=softmax(\frac{\textbf{D}^{x}_k\cdot\textbf{Q}\cdot \textbf{D}^{y}_k}{\sqrt{d}})
\end{split}
\end{equation}
where $\textbf{P}$ and $\textbf{Q}$ are parameter matrices trained with the whole network. The $softmax(\cdot)$ operation is applied to each row vector of its input matrix, and $d$ is the column number of  $\textbf{P}$ and $\textbf{Q}$. Following Vaswani ~\emph{et al.} \cite{transformer}, $\frac{1}{\sqrt{d}}$ is used as a normalization factor to keep $softmax(\cdot)$  from the region with an extremely small gradient. Based on the affinity matrices, $\textbf{D}^x_k$ and $\textbf{D}^y_k$ are fused to capture the joint distribution on the $x$ and $y$ axes, as shown in Equation \eqref{att2}:

\begin{equation}
\begin{split}
\label{att2}
\textbf{D}^{x'}_{k}=\textbf{D}^{x}_k+\gamma\textbf{W}_k^{yx}\cdot(\textbf{D}^{y}_k)^\mathrm{T},\,\,\,\,
\textbf{D}^{y'}_{k}=\textbf{D}^{y}_k+\gamma(\textbf{W}_k^{xy}\cdot\textbf{D}^{x}_k)^\mathrm{T}
\end{split}
\end{equation}
where $\gamma$ is the weight of the attentive feature. Next, $\textbf{D}^{x'}_k$ is fed into $CNN^x_2$ as an input channel, and $\textbf{D}^{y'}_k$ is fed into $CNN^y_2$.

\subsubsection{Loss Function}
$g_{de}(\cdot)$ is trained by supervised regression. The error between the prediction and ground truth is minimized as shown in Equation \eqref{loss1}. $L_{de}$ is optimized by an Adam optimizer with a learning rate of 1e-4.
\begin{equation}
\label{loss1}
\begin{split}
\min_{\theta_{de}} L_{de}&=\sum_{n=1}^N(||\textbf{h}^x_n-\textbf{h}^{x*}_{n}||_2^2+||\textbf{h}^y_n-\textbf{h}^{y*}_{n}||_2^2)\\
&=\sum_{n=1}^N(||g_{de}(\textbf{I},n,x;\theta_{de})-\textbf{h}^{x*}_{n}||_2^2+||g_{de}(\textbf{I},n,y;\theta_{de})-\textbf{h}^{y*}_{n}||_2^2)
\end{split}
\end{equation}
where $\textbf{h}^{x*}_{n}$ and $\textbf{h}^{y*}_{n}$ are the ground truths. They are the marginal distributions of $\textbf{H}^*_{n}(x,y)$, as shown in Equation \eqref{marginal}:
\begin{equation}
\begin{split}
\label{marginal}
\textbf{h}^{x*}_{n}=\sum_y\textbf{H}^*_{n}(x,y),\,\,\,
\textbf{h}^{y*}_{n}= \sum_x\textbf{H}^*_{n}(x,y)
\end{split}
\end{equation} 

\subsection{Tracker}
\subsubsection{Integrating Spatial and Temporal Patterns} First, spatial patterns on the current frame are encoded by the proposed detector, as shown in the left part of Fig. \ref{tracker}. The detected heatmaps on the $x$ and $y$ axes are stacked as matrices, denoted as $\textbf{E}_t^x$ and $\textbf{E}_t^y$, respectively, for the $t$ th frame. 

Second, the detected heatmaps are refined by temporal patterns. This is beneficial because the facial appearance on the current frame may not be reliable due to some ``in the wild" disturbances, such as occlusions or uneven illuminations. Integrating temporal patterns from previous frames may help locate landmarks when spatial features on the current frame are unreliable. As depicted in the green border boxes of Fig. \ref{tracker}, $CNN_3^x$ and $CNN_3^y$ encode $\textbf{E}_t^x$ and $\textbf{E}_t^y$ as features, denoted as $\textbf{U}_t^x$ and $\textbf{U}_t^y$, respectively. In the feature space, $\textbf{U}_t^x$ and $\textbf{U}_t^y$ are fused with features from the past frames, as shown in Equation \eqref{temporal}:

\begin{equation}
\label{temporal}
\textbf{V}_t^x=\textbf{U}_t^x+\sum_{\tau=1}^{t-1}\lambda^{t-\tau}\textbf{U}_\tau^x,\,\,\,
\textbf{V}_t^y=\textbf{U}_t^y+ \sum_{\tau=1}^{t-1}\lambda^{t-\tau}\textbf{U}_\tau^y
\end{equation} 
where $\textbf{V}_t^x$ and $\textbf{V}_t^y$ are features embedded with temporal patterns, $\lambda\in[0,1]$ is a hyper-parameter to attenuate the weights of frames far from the current. To generate heatmap refinements, $CNN_4^x$ decodes $\textbf{V}_t^x$ to $\textbf{E}_t^{x'}$ and $CNN_4^y$ decodes $\textbf{V}_t^y$ to $\textbf{E}_{t}^{y'}$. $\textbf{E}_t^{x'}$ and $\textbf{E}_{t}^{y'}$ are used to refine  $\textbf{E}_t^x$ and $\textbf{E}_{t}^y$ respectively by adding together with them as the tracking results.

\subsubsection{Loss Function}
Similar to the detector, the tracker $g_{tr}$ is also trained by supervised regression. The training loss is shown in Equation \eqref{loss2}:
\begin{equation}
\label{loss2}
\begin{split}
\min_{\theta_{tr}} L_{tr}=&\sum_{t=1}^T\sum_{n=1}^N(||\textbf{h}^x_{tn}-\textbf{h}^{x*}_{tn}||_2^2+||\textbf{h}^y_{tn}-\textbf{h}^{y*}_{tn}||_2^2)\\
=&\sum_{t=1}^T\sum_{n=1}^N(||g_{tr}(\textbf{I}_{1:t},n,x;\theta_{tr})-\textbf{h}^{x*}_{tn}||_2^2\\&+||g_{tr}(\textbf{I}_{1:t},n,y;\theta_{tr})-\textbf{h}^{y*}_{tn}||_2^2)
\end{split}
\end{equation}
\section{Experiments}

\subsection{Experimental Conditions}
Facial landmark detection experiments are conducted on the 300W \cite{sagonas2016300} and the AFLW dataset, both of them are image datasets.
Facial landmark tracking experiments are conducted on the
300VW \cite{shen2015first} and the Talking Face (TF) \cite{FGNET} dataset, they are video datasets. 

The 300W dataset contains 68 pre-defined landmarks. For experiments on the 300W dataset, the proposed method is trained on its training set with $3,148$ images, and evaluated on its public testing set, composed of a common subset with $554$ images and a challenging subset with $135$ images.  For saving space, in the following part of the paper, their names are simplified as 300W \textit{com} and \textit{cha}, respectively. The full public testing set is  simplified as 300W \textit{full}.

The AFLW dataset has 21 pre-defined landmarks. For experiments on the AFLW dataset, we just use 19 landmarks as previous work did \cite{Dong_2019_ICCV}. The proposed method is trained on the training set with 20000 images, and evaluated on the testing set with 4386 images. 

The 300VW dataset contains 68 pre-defined landmarks. It has a training set with 50 videos, and a testing set consisting of 60 videos from three difficulty levels, i.e., well-lit (scenario 1), mild unconstrained (scenario 2) and challenging (scenario 3). Their names are simplified as 300VW S1, S2, and S3, respectively. For experiments on the 300VW dataset, following Yin ~\emph{et al.} \cite{GAN_Tracking}, our method is trained on the union of training sets from the 300VW and 300W dataset, and evaluated on the 300VW testing set. Since the 300W dataset only contains images with no temporal information, for the tracking task, we only use it to train the detector inside the tracker.

Since the TF dataset only contains one video with 5000 frames, the method is trained on the 300VW dataset and evaluated on the TF dataset. Due to the different landmark definitions between the TF and the 300VW dataset, we follow Liu ~\emph{et al.} \cite{TSTN} to apply the seven common landmarks for testing. 
\begin{table*}[t]
	\centering
	
	\begin{tabular}{c |c |c c c| c  c c c c c c} \hline
			\multicolumn{2}{c|}{$L$} &64 &128&256&64 &128&256&384&512&640&768\\\hline
			\multicolumn{2}{c|}{$L/F$} & 0.25 &0.5 & 1.0&0.25 & 0.5&1.0 &1.5 &2.0 &2.5 &3.0 \\\hline\hline
			\multicolumn{2}{c|}{Method} &\multicolumn{3}{c|}{2D heatmap-based detector}&\multicolumn{7}{c}{The proposed 1D heatmap-based detector}\\\hline
			\multirow{3}{*}{\rotatebox{90}{NRMSE}}&300W \textit{com}&3.53&3.26&OOM &3.45&3.22&3.10& 3.01 &2.97&2.93 &\textbf{2.91} \\\cline{2-12} 
			&300W \textit{cha}&6.16&5.62&OOM&6.22 &5.68 & 5.45& 5.40&5.39&5.35 &\textbf{5.31} \\\cline{2-12}
			&300W \textit{full} &4.04&3.72&OOM&3.99&3.70 &3.56 &3.48 &3.44 & 3.40 &\textbf{3.38}
			\\\hline\hline
			\multicolumn{2}{c|}{Method} &\multicolumn{3}{c|}{2D heatmap-based tracker}&\multicolumn{7}{c}{The proposed 1D heatmap-based tracker}\\\hline
			\multirow{3}{*}{\rotatebox{90}{NRMSE}}&300VW S1 &4.53&4.27&OOM &3.61&3.47 &3.37 &3.29 &3.20&3.12 & \textbf{3.06} \\\cline{2-12}
			&300VW S2 &4.60&4.34&OOM&3.90&3.71 &3.54&3.43 &3.33&3.24 &\textbf{3.17} \\\cline{2-12}
			&300VW S3& 5.97&5.72&OOM & 4.75&4.52 &4.39 &4.32&4.24&4.18 & \textbf{4.12} \\\hline
	\end{tabular}
	
	\caption{NRMSE (\%) of the 2D heatmap-based detector, the proposed detector, the 2D heatmap-based tracker and the proposed tracker with different output resolutions ($L$).The input face resolution (F) is fixed as 256. OOM is the abbreviation of ``Out of Memory".}
	\label{resolution_all}
\end{table*}

The detector and tracker are evaluated by the accuracy of their predictions, which is quantified by the \textit{N}ormalized \textit{R}oot \textit{M}ean \textit{S}quared \textit{E}rror (NRMSE) between the predicted landmark coordinates and the ground truths. A lower NRMSE corresponds to a better accuracy. Following previous works \cite{sagonas2016300,Dong_2019_ICCV,GAN_Tracking}, on the 300W, the 300VW, and the TF datasets, the error is normalized by the inter-ocular distance of a face. On the AFLW dataset,  it is normalized by the face size.  

All experiments are conducted by Tensorflow 1.9.0 on a NVIDIA TESLA V100 GPU with 32GiB memory. A face is firstly cropped from the bounding box and scaled to $256\times256$ pixels, then fed into the detector or the tracker. A training batch contains $10$ facial images or frames. The resolution of the 1D heatmap is set to three times the face size. For the experiment on each dataset, the training set is splitted as 10 folds to conduct cross validation and select optimal values for $\gamma$ in Equation \eqref{att2} and $\lambda$ in Equation \eqref{temporal}. $\gamma$ and $\lambda$ are searched from $\{0.1,0.2,...,1.0\}$. The optimal $\gamma$ found on the 300W training set, the AFLW training set, and the 300VW training set are $0.4$, $0.5$ and $0.4$, respectively. The optimal $\lambda$ found on the 300VW training set is $0.3$.

In the following parts, we make quantitative evaluation on the proposed detector and tracker under different parameter settings, and compare them with related works. All hyper-parameters are assigned with their optimal values except for the parameter we want to further study on.

\subsection{Experimental Results and Analysis under Different
	Heatmap Resolutions}
Table \ref{resolution_all} shows the NRMSE performance of the proposed detector and tracker with different heatmap resolutions. Due to the space limitation of the paper, Table \ref{resolution_all} only displays the results on the 300W and the 300VW datasets, which have the most pre-defined landmarks. To compare with 2D heatmaps, we also implement a 2D heatmap-based detector and tracker and display their performance in Table \ref{resolution_all}. The 2D heatmap-based detector is the same as FAN \cite{how_far}, which takes the output of the stacked hourglass network as 2D heatmaps. Based on the 2D heatmaps predicted by the detector, the temporal recurrent learning method proposed by Xi ~\emph{et al.} \cite{encoder_decoder} is adopted as the compared tracker.

From Table \ref{resolution_all}, we have the following observations. First, when $L$ increases to $256$, the 2D heatmap-based methods crash by memory overload, which means the memory requirement is larger than the capacity (32 GiB) of the GPU. The huge output complexity of 2D heatmaps restricts $L$ to a low value, i.e., $128$, which is lower than the face size and causes great quantization errors. Second, compared to the 2D heatmap regression methods, the proposed method can achieve a much larger output resolution under limited GPU memory because the low spatial complexity ($2NL$) of 1D heatmap does not take significant memory overhead. Third, as $L$ increases, NRMSE decreases. This is because with the increase of $L$, the quantization error is reduced  and more detailed spatial and temporal patterns are captured. For example, as $L$ increases from $64$ to $768$, the NRMSE of our method decreases by $15.65\%$, $14.63\%$, and $15.28\%$ on the 300W \textit{com}, \textit{cha}, and \textit{full}. The light-weight 1D heatmap allows us to fully boost its resolution despite limited space, and the quantization error is well alleviated. With a much higher output resolution, the proposed detector and tracker outperform the 2D heatmap-based detector and tracker significantly on accuracy. Fourth, the proposed 1D heatmap-based tracker outperforms the 2D heatmap-based tracker significantly even if the output resolution is the same. The reason may be that, it is hard for the tracker to process  the huge structure of the temporal sequence of 2D heatmaps. The proposed tracker solves the problem by only processing the light-weight 1D heatmaps, and therefore the performance is much better.

\begin{figure}[t]
	\centering
	\begin{minipage}{0.48\linewidth}
		\centering
		\includegraphics[width = 40 mm, height = 11mm]{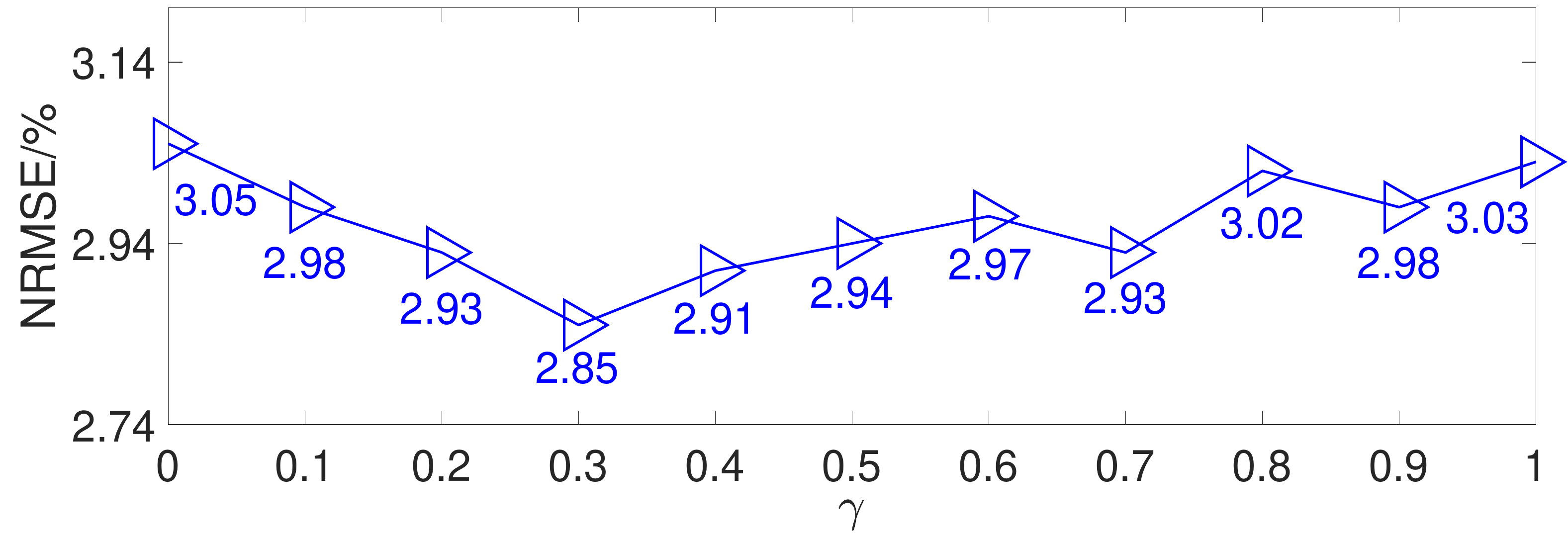}
		\subcaption{}
		\label{300W_com}
	\end{minipage}%
	\begin{minipage}{0.48\linewidth}
		\centering
		\includegraphics[width = 40 mm, height = 11mm]{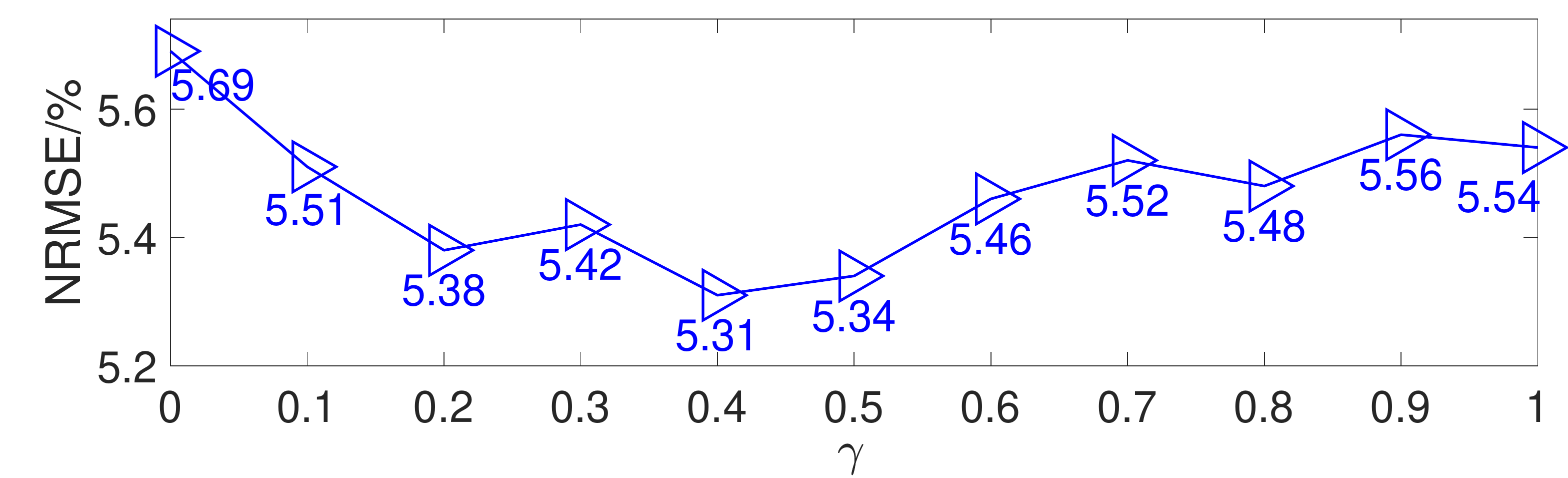}
		\subcaption{}
		\label{300W_cha}
	\end{minipage}
	
	\begin{minipage}{0.48\linewidth}
		\centering
		\includegraphics[width = 40 mm, height = 11mm]{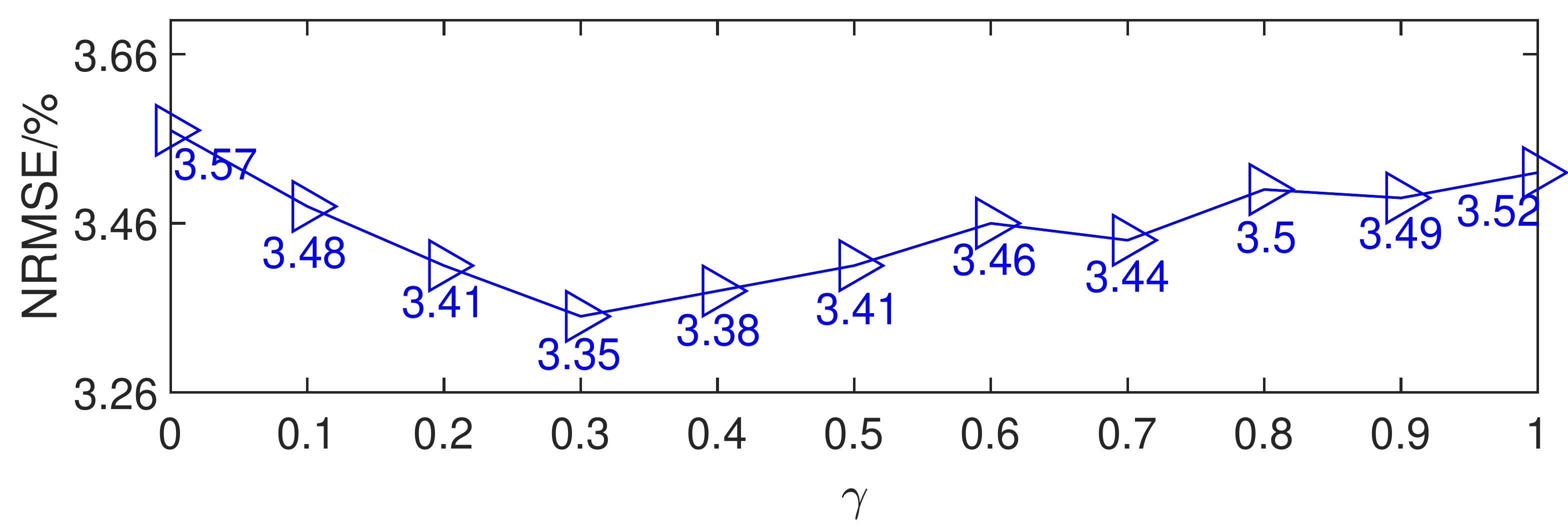}
		\subcaption{}
		\label{300W_all}
	\end{minipage}
	\begin{minipage}{0.48\linewidth}
		\centering
		\includegraphics[width = 40 mm, height = 11mm]{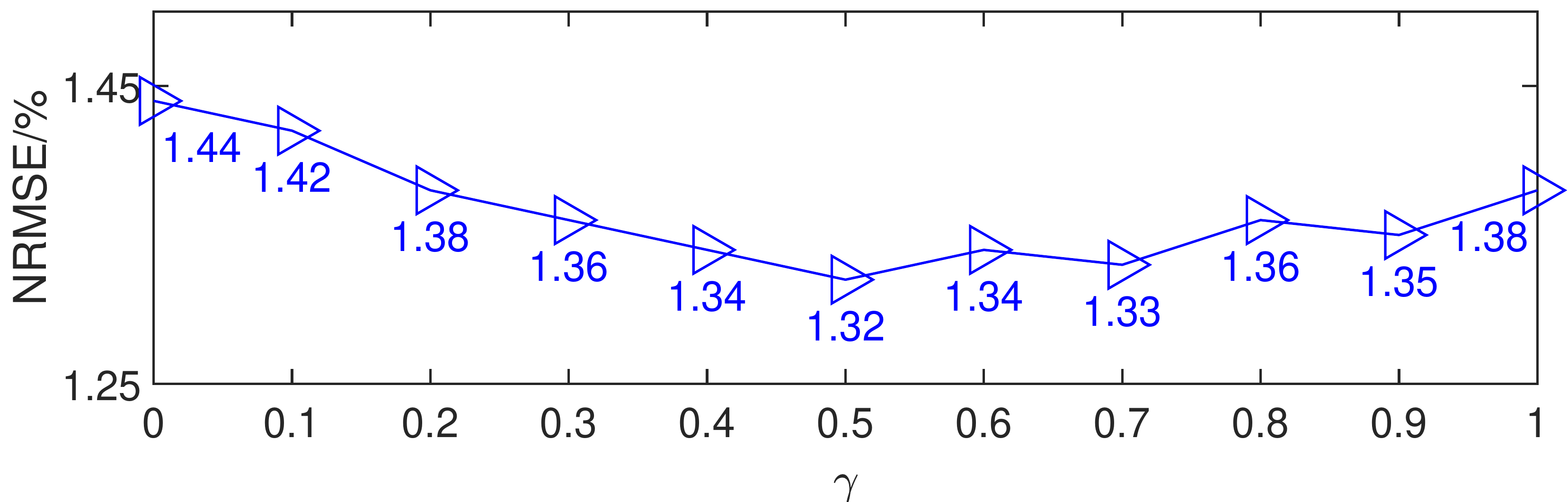}
		\subcaption{}
		\label{AFLW}
	\end{minipage}%
	
	\begin{minipage}{0.48\linewidth}
		\centering
		\includegraphics[width = 40 mm, height = 11mm]{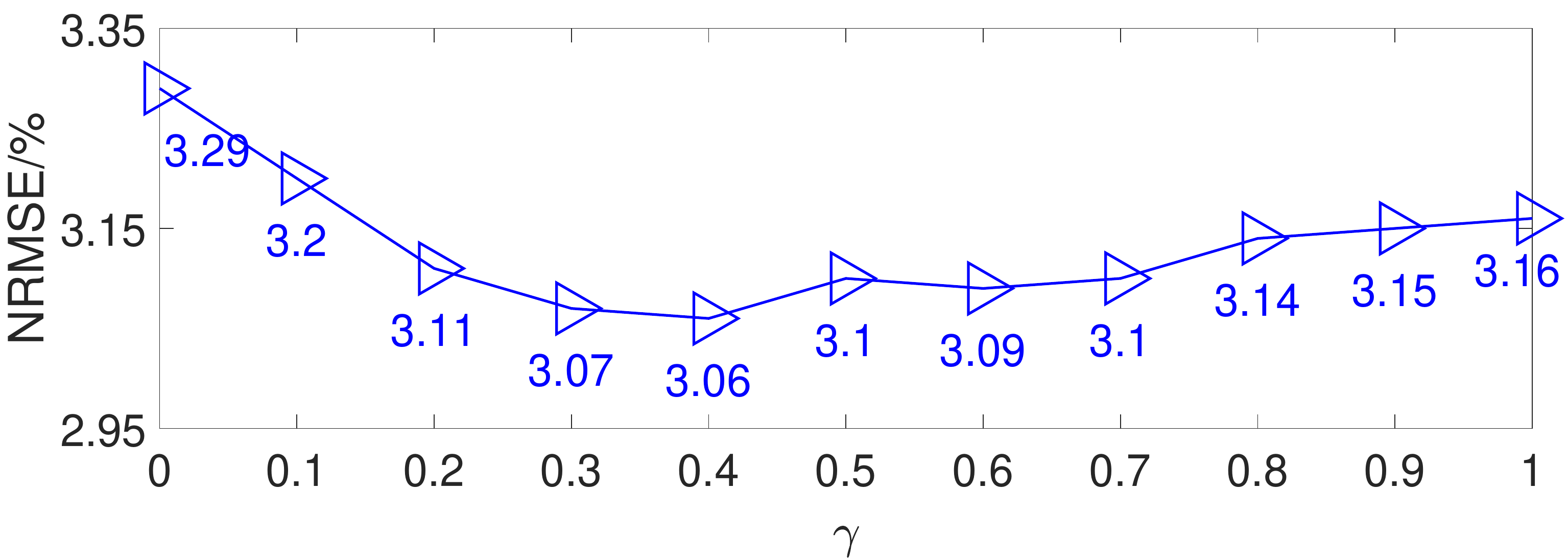}
		\subcaption{}
		\label{300VW_S1}
	\end{minipage}%
	\begin{minipage}{0.48\linewidth}
		\centering
		\includegraphics[width = 40 mm, height = 11mm]{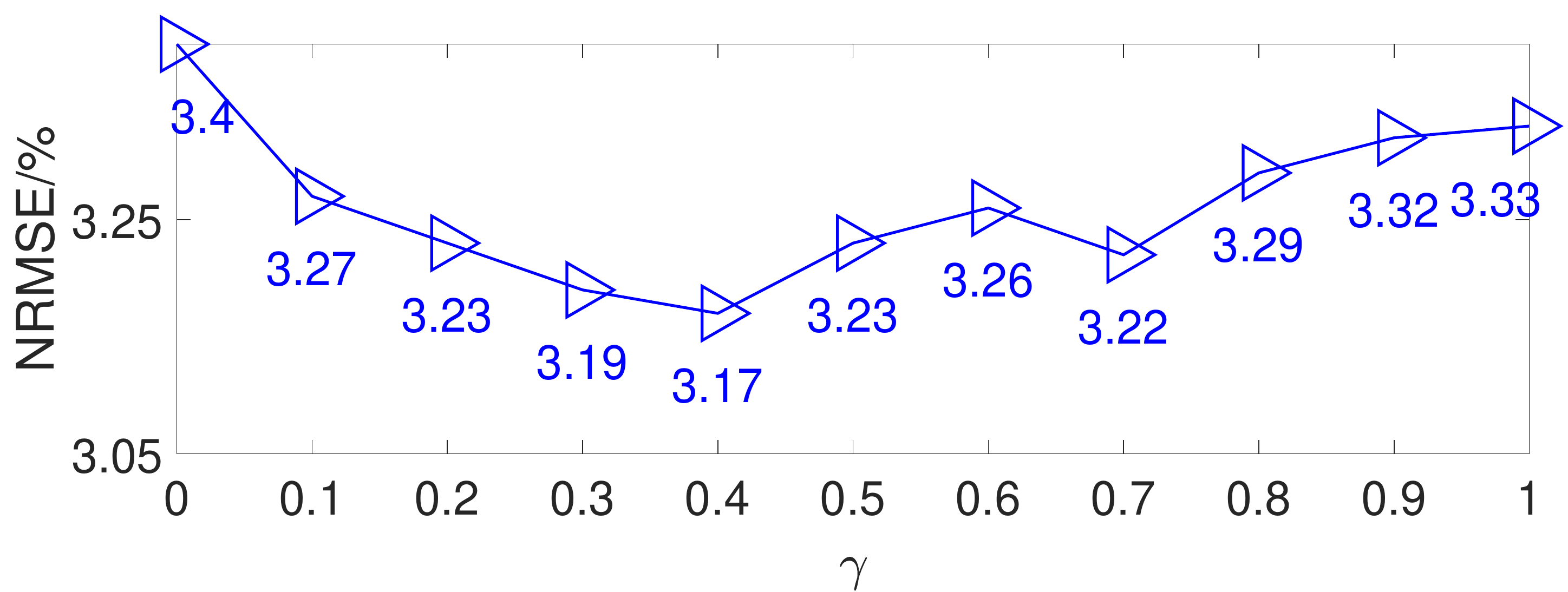}
		\subcaption{}
		\label{300VW_S2}
	\end{minipage}
	
	\begin{minipage}{0.48\linewidth}
		\centering
		\includegraphics[width = 40 mm, height = 11mm]{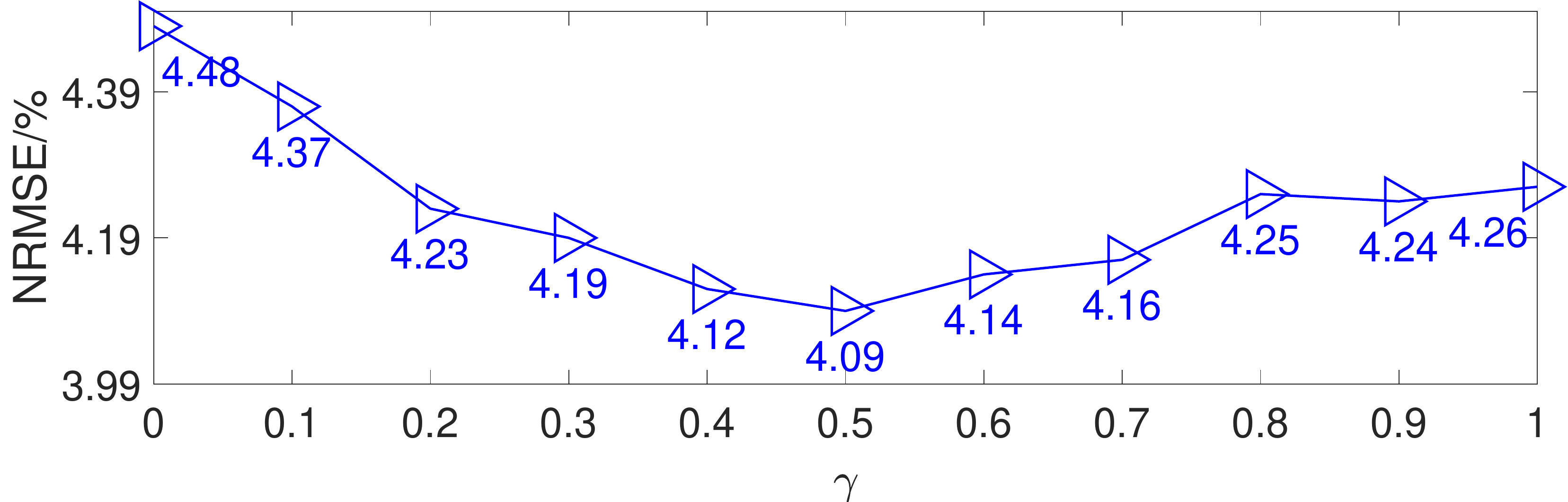}
		\subcaption{}
		\label{300VW_S3}
	\end{minipage}
	\begin{minipage}{0.48\linewidth}
		\centering
		\includegraphics[width = 42 mm, height = 11mm]{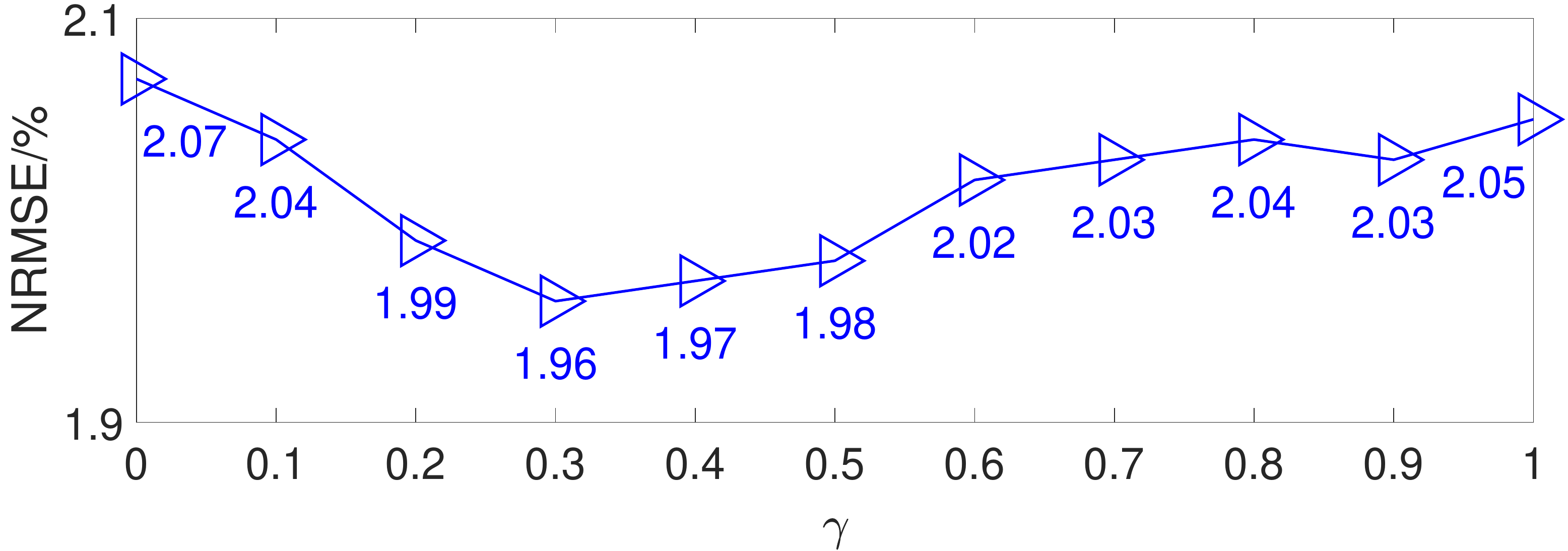}
		\subcaption{}
		\label{TF}
	\end{minipage}
	
	\caption{NRMSE (\%) performance on (a) 300W \textit{com}, (b) 300W \textit{cha}, (c) 300W \textit{full}, (d) AFLW, (e) 300VW S1, (f) 300VW S2, (g) 300VW S3 and (h) TF with different $\gamma$.}
	\label{gamma}
\end{figure}

\subsection{Ablation Study for the Co-Attention Module}
The proposed co-attention module shares distributional features between the $x$ and $y$ axes to implicitly capture joint distributions. According to Equation \eqref{att2}, the weight of the attentive feature is controlled by the parameter $\gamma$. When $\gamma$ is $0.0$, the co-attention module is discarded and the method only models marginal distributions separately on the two axes. As $\gamma$ increases, the weight of the attentive feature in the fused representations $\textbf{D}^{x'}_k$ and $\textbf{D}^{y'}_k$ grows. We display  NRMSE performance with different $\gamma$ values in Fig. \ref{gamma}. 

From Fig. \ref{gamma}, we find that there is a significant boost on detecting and tracking accuracy when $\gamma$ increases from 0.0 to 0.4. Specifically, NRMSE decreases by $4.59\%$, $6.68\%$ and  $5.32\%$  on 300W \textit{com}, \textit{cha} and \textit{full}, respectively. It also decreases by $6.94\%$ on the AFLW dataset; and by $6.99\%$, $6.76\%$ and $8.04\%$ on the three scenarios of the 300VW dataset; and by $4.83\%$ on the TF dataset. That demonstrates the effectiveness of the co-attention module.

\subsection{Ablation Study for the Temporal Refinement Mechanism of the Tracker}
The tracker refines the heatmaps predicted by the detector by integrating temporal patterns from past frames, as shown in Equation \eqref{temporal}. The weight of features from past frames is determined by the parameter $\lambda$. We make ablation study for the temporal refinement mechanism by comparing the results of two experimental settings. For the first setting, temporal refinement is discarded by assigning $\lambda$ as $0.0$. For the second setting, temporal refinement is kept by assigning $\lambda$ as $0.3$, the optimal value found by cross validation. Results of the two settings are shown in Table \ref{abla_tracker1} and splitted by a slash. From Table \ref{abla_tracker1}, we find that the temporal refinement mechanism is beneficial to tracking accuracy. When $\lambda$ is $0.3$, NRMSE decreases  by $7.55\%$, $8.12\%$, $6.79\%$ and $2.48\%$ on the three scenarios of the 300VW dataset and the TF datasets, respectively. This comparison demonstrates the effectiveness of the temporal refinement mechanism.

\begin{table}
	\centering
	\setlength{\tabcolsep}{1.3mm}{\begin{tabular}{c | c c c c} \hline
			Dataset&300VW S1 &300VW S2&300VW S3&TF\\\hline
			NRMSE&3.31/\textbf{3.06} &3.45/\textbf{3.17} &4.42/\textbf{4.12} &2.02/\textbf{1.97}\\\hline
	\end{tabular}}
	\caption{NRMSE (\%) of the proposed tracker without/with temporal refinement.}
	\label{abla_tracker1}
\end{table}

\begin{table*}
	\centering
    \begin{tabular}{c |c c c c c c c cc c  c c|| c  c} \hline
			Method&SAN&CNN-CRF&DSRN&LAB&LaplaceKL &ODN& \cite{better_heatmap1}&  DSNT & DARK &FHR&\cite{chen2019adversarial}&\cite{hourglass}&Ours&Ours+DARK\\\hline
			300W com &3.34 &3.33&4.12&2.98 &3.19  & 3.56& 3.22&3.17 &2.98 &3.04 &- &3.30  &2.91&\textbf{2.68}\\\hline
			300W cha  &6.60 &6.29&9.68&5.19&6.87 &6.67&  5.62 &5.54 &5.48 &6.21&-&5.69 &5.31 &\textbf{4.92}\\\hline
			300W full &3.98 & 3.91&5.21&3.49 & 3.91&4.17& 3.69 &3.63 &3.47&3.66&- &3.77&3.38 &\textbf{3.12}
			\\\hline
			AFLW&1.91 &-&1.86&1.25 &1.97 & 1.63& 1.80 & 1.74&1.43 &1.58 &1.39&1.95&1.32 &\textbf{1.17}			 \\\hline
	\end{tabular}
	\caption{NRSME (\%) of the proposed detector and the compared methods on the image datasets.}
	\label{res_300w}
\end{table*}

\begin{table*}
	\centering
	\begin{tabular}{c | c c c c c c c c  c c c c || c} \hline
		Method&SDM &TSCN& CFAN  & CFSS& IFA&TCDCN&TSTN&GHCU&\cite{Sun_2019_ICCV}&FHR &FHR+STA& GAN& Ours\\\hline
		300VW S1 &7.41 &12.54  & - &7.68& - &7.66 & 5.36 &3.85&3.56&4.82 &4.21 &3.50&\textbf{3.06} \\\hline
		300VW S2 &6.18   &7.25 & -&6.42&- &6.77 &4.51 &3.46&3.88& 4.23 &4.02 &3.67&\textbf{3.17}\\\hline
		300VW S3 & 13.04 & 13.13 & - & 13.67&-& 14.98&  12.84 & 7.51&5.02&7.09 & 5.64&4.43&\textbf{4.12}\\\hline
		TF& 4.01 & - & 3.52 & 2.36& 3.45 &- & 2.13&- &- &2.07&2.10&2.03&\textbf{1.97}\\\hline 
	\end{tabular}
	\caption{NRSME (\%) of the proposed tracker and the compared methods on the 300VW and the TF datasets}
	\label{res_300vw}
\end{table*}

\subsection{Comparison with State-of-the-art Methods}

The proposed method is compared to other state-of-the-art landmark localization methods. From these approaches, coordinate regression methods include SDM \cite{SDM}, TSCN \cite{TSCN}, IFA \cite{IFA},  CFSS \cite{CFSS}, TCDCN \cite{TCDCN}, TSTN \cite{TSTN}, DSRN \cite{DSRN}, ODN \cite{Zhu_2019_CVPR}, STA \cite{tai2018towards}, Sun~\emph{et al.}'s work \cite{Sun_2019_ICCV} and GAN \cite{GAN_Tracking}. Heatmap regression methods include Newell ~\emph{et al.}'s work \cite{hourglass}, SAN \cite{SAN}, LAB \cite{LAB}, CNN-CRF \cite{CNNCRF}, LaplaceKL \cite{Robinson_2019_ICCV}, Sun~\emph{et al.}'s work \cite{better_heatmap1}, DSNT \cite{better_heatmap2} , DARK \cite{better_heatmap3}, FHR \cite{tai2018towards}, GHCU \cite{Liu_2019_CVPR} and Chen~\emph{et al.}'s work \cite{chen2019adversarial}. From these methods, CFSS, TCDCN, DSRN, ODN, Newell ~\emph{et al.}'s work, SAN, LaplaceKL, Sun~\emph{et al.}'s work \cite{better_heatmap1}, DSNT, DARK, FHR, GHCU, Chen~\emph{et al.}'s work, LAB and CNN-CRF are detection methods. TSCN, TSTN, STA, Sun~\emph{et al.}'s work \cite{Sun_2019_ICCV} and GAN are tracking methods. SDM, IFA contain both a detection method and a tracking method.  Some newly proposed semi-supervised learning or unsupervised learning  methods \cite{SBR,Dong_2019_ICCV,Qian_2019_ICCV} for landmark detection are trained under different conditions with our method so are not included in our comparison.

Table \ref{res_300w} lists the NRSME performance of the proposed detector and the compared methods on the image datasets, i.e., the 300W public testing set and the AFLW dataset. Table \ref{res_300vw} lists the NRSME performance of the proposed tracker and the compared methods on the video datasets, i.e., the 300VW and the TF dataset, respectively. Performances of the compared methods are directly copied from literature, except for that of Sun~\emph{et al.}'s work \cite{better_heatmap1}, DSNT, DARK, FHR in Table \ref{res_300w} because we could not find their published results on the respective datasets. We just re-implement them under the same experimental condition with our method. Their open source codes\footnote{https://github.com/JimmySuen/integral-human-pose\\ https://github.com/anibali/dsntnn\\ https://ilovepose.github.io/coco/\\
	https://github.com/tyshiwo/FHR\_alignment} are used to facilitate re-implementation. Other methods lacking published results or evaluated under different metrics or normalization standards are just left blank.

From Table \ref{res_300w}, we find that our method achieves state-of-the-art detecting accuracy on the image datasets. We outperform Sun~\emph{et al.}'s work \cite{better_heatmap1}, DSNT, DARK and FHR, which also try to capture the fractional parts of coordinates and reduce quantization errors. These methods still maintain a 2D heatmap structure whose resolution is limited by the memory capacity, and therefore suffering from the information loss caused by the quantization process. Our method significantly boosts the output resolution with a light-weight output structure, i.e., the 1D heatmap, which captures more detailed distributional information and significantly alleviates the quantization error. We also combine our method with DARK by adapting DARK to 1D Gaussian distribution to estimate ground truth positions from the predicted 1D heatmaps. As shown in Table \ref{res_300w}, the result of Ours+DARK outperforms the original DARK. This result further demonstrates the superiority of the proposed 1D heatmap structure.
\begin{figure*}
	\centering
	\begin{minipage}[t]{0.497\linewidth}
		\centering
		\includegraphics[width = 65 mm, height = 23mm]{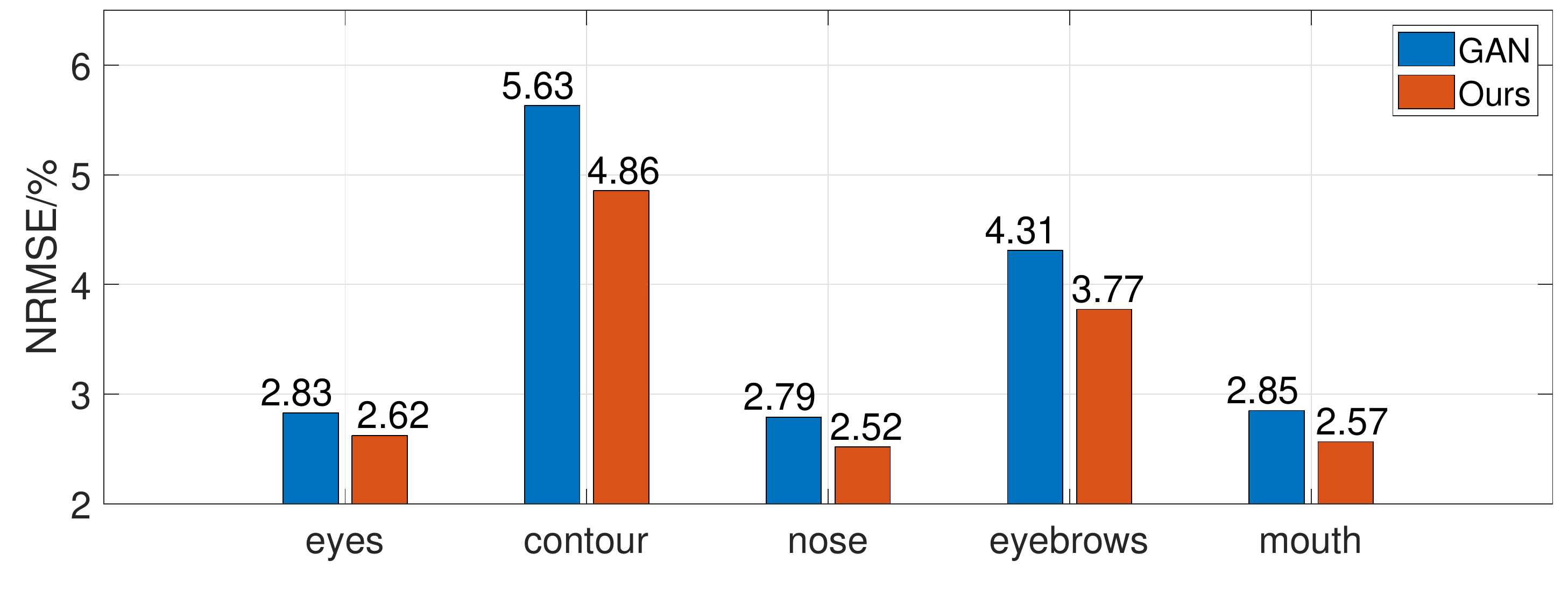}
		\caption{NRMSE (\%) performance on each facial area.}
		\label{dis1}
	\end{minipage}
	\begin{minipage}[t]{0.497\linewidth}
		\centering
		\includegraphics[width = 65 mm, height = 23mm]{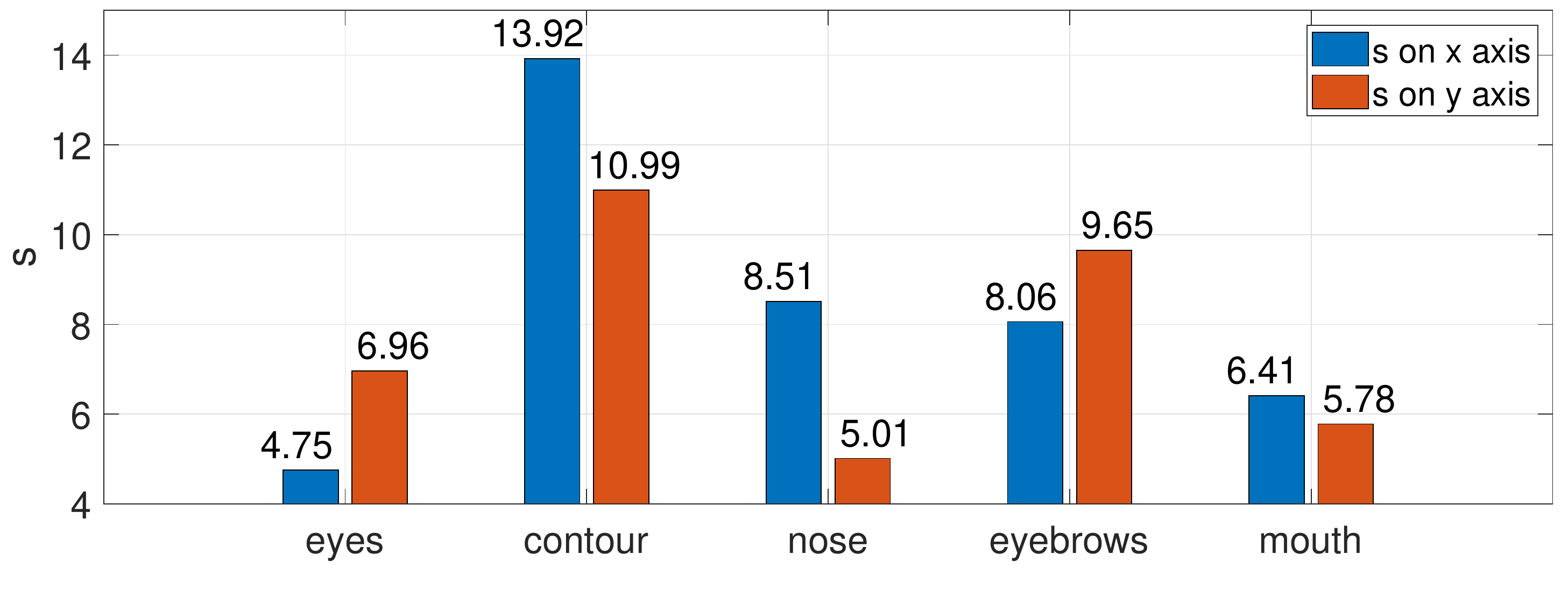}
		\caption{Average value ($s$) of coordinate variances within each facial area.}
		\label{dis2}
	\end{minipage}
\end{figure*} 

\begin{figure*}
	\centering
	\begin{minipage}[t]{0.078\linewidth}
		\centering
		\includegraphics[scale=0.15]{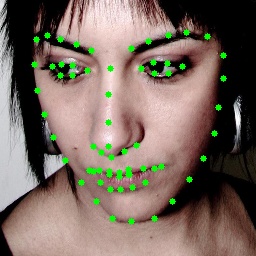}
		
	\end{minipage}
	\begin{minipage}[t]{0.078\linewidth}
		\centering
		\includegraphics[scale=0.15]{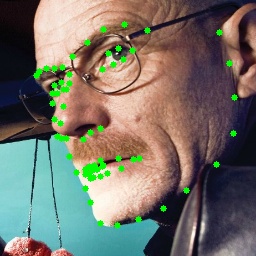}
		
		\label{h1}
	\end{minipage}
	\begin{minipage}[t]{0.078\linewidth}
		\centering
		\includegraphics[scale=0.15]{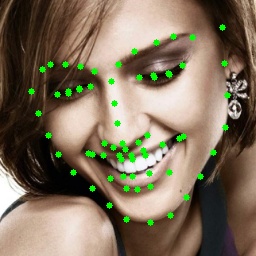}
		
	\end{minipage}
	\begin{minipage}[t]{0.078\linewidth}
		\centering
		\includegraphics[scale=0.15]{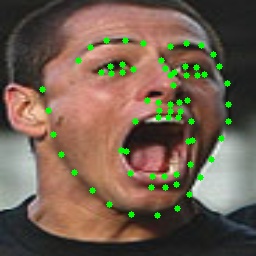}
		
		\label{d1}
	\end{minipage}
	\begin{minipage}[t]{0.078\linewidth}
		\centering
		\includegraphics[scale=0.15]{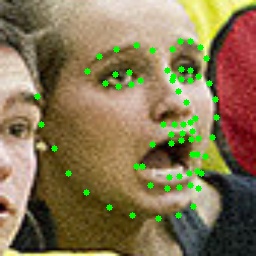}
	\end{minipage}
	\begin{minipage}[t]{0.078\linewidth}
		\centering
		\includegraphics[scale=0.15]{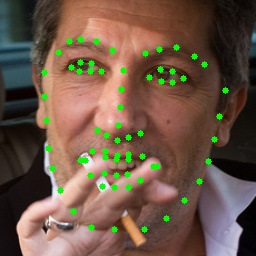}
	\end{minipage}
	\begin{minipage}[t]{0.078\linewidth}
		\centering
		\includegraphics[scale=0.15]{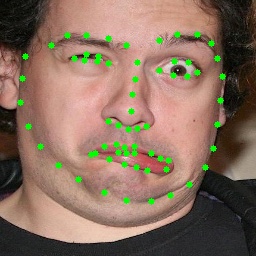}
		\subcaption{}
	\end{minipage}
	\begin{minipage}[t]{0.078\linewidth}
		\centering
		\includegraphics[scale=0.15]{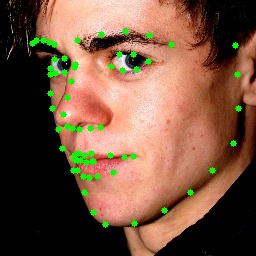}
	\end{minipage}
	\begin{minipage}[t]{0.078\linewidth}
		\centering
		\includegraphics[scale=0.15]{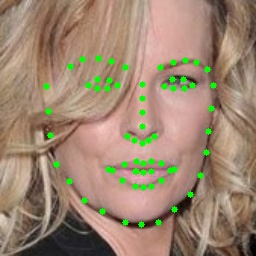}
	\end{minipage}
	\begin{minipage}[t]{0.078\linewidth}
		\centering
		\includegraphics[scale=0.15]{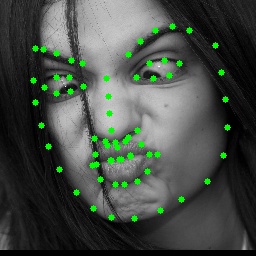}
	\end{minipage}
	\begin{minipage}[t]{0.078\linewidth}
		\centering
		\includegraphics[scale=0.15]{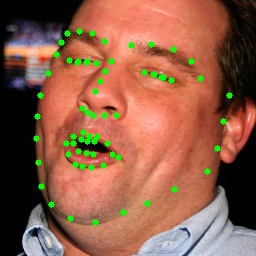}
	\end{minipage}
	\begin{minipage}[t]{0.078\linewidth}
		\centering
		\includegraphics[scale=0.15]{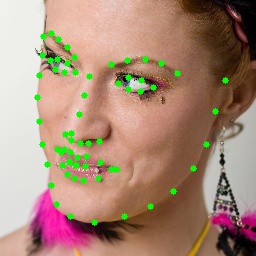}
	\end{minipage}
	
	\begin{minipage}[t]{0.078\linewidth}
		\centering
		\includegraphics[scale=0.15]{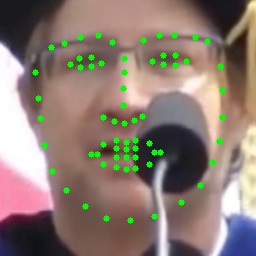}
	\end{minipage}
	\begin{minipage}[t]{0.078\linewidth}
		\centering
		\includegraphics[scale=0.15]{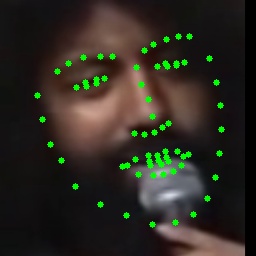}
	\end{minipage}
	\begin{minipage}[t]{0.078\linewidth}
		\centering
		\includegraphics[scale=0.15]{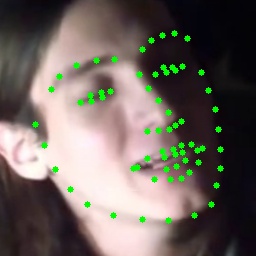}
	\end{minipage}
	\begin{minipage}[t]{0.078\linewidth}
		\centering
		\includegraphics[scale=0.15]{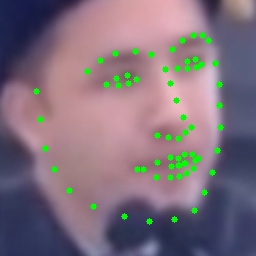}
	\end{minipage}
	\begin{minipage}[t]{0.078\linewidth}
		\centering
		\includegraphics[scale=0.15]{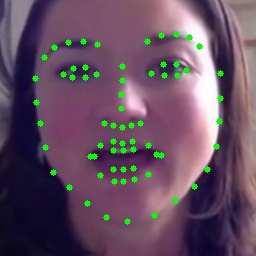}
	\end{minipage}
	\begin{minipage}[t]{0.078\linewidth}
		\centering
		\includegraphics[scale=0.15]{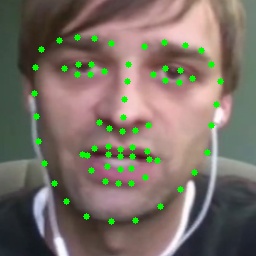}
	\end{minipage}
	\begin{minipage}[t]{0.078\linewidth}
		\centering
		\includegraphics[scale=0.15]{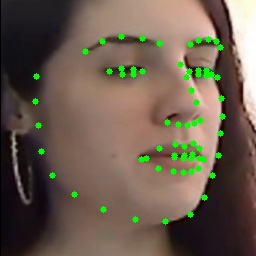}
		\subcaption{}
	\end{minipage}
	\begin{minipage}[t]{0.078\linewidth}
		\centering
		\includegraphics[scale=0.15]{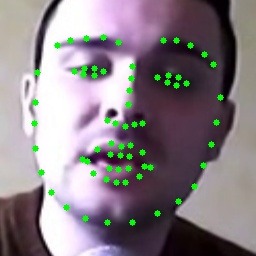}
	\end{minipage}
	\begin{minipage}[t]{0.078\linewidth}
		\centering
		\includegraphics[scale=0.07625]{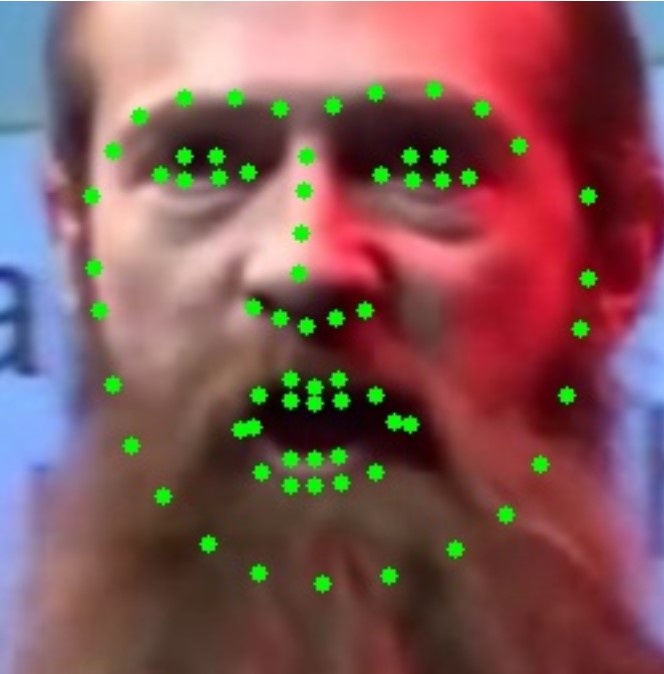}
	\end{minipage}
	\begin{minipage}[t]{0.078\linewidth}
		\centering
		\includegraphics[scale=0.15]{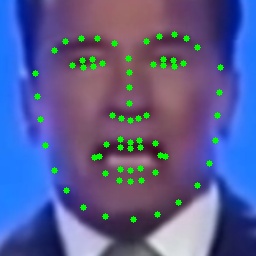}
	\end{minipage}
	\begin{minipage}[t]{0.078\linewidth}
		\centering
		\includegraphics[scale=0.1025]{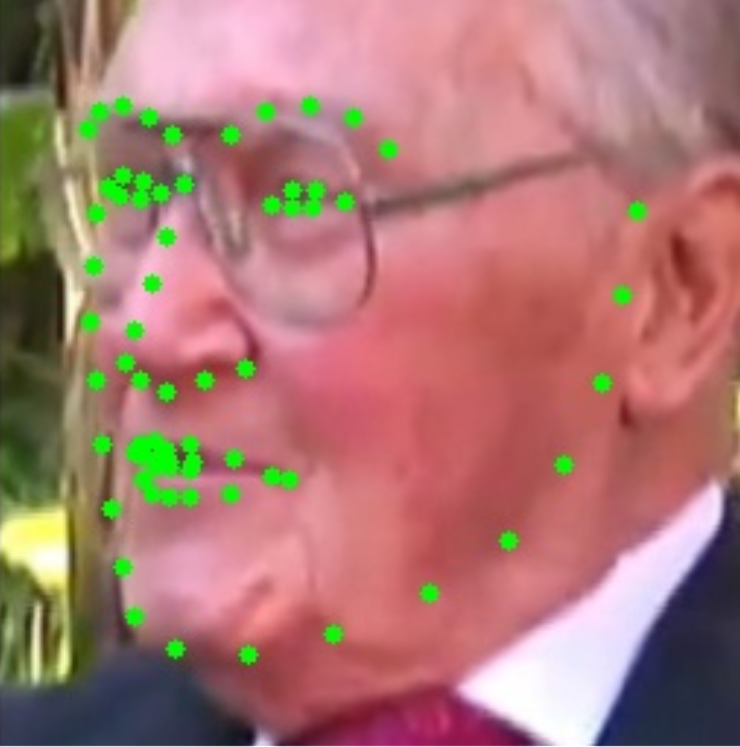}
	\end{minipage}
	\begin{minipage}[t]{0.078\linewidth}
		\centering
		\includegraphics[scale=0.093]{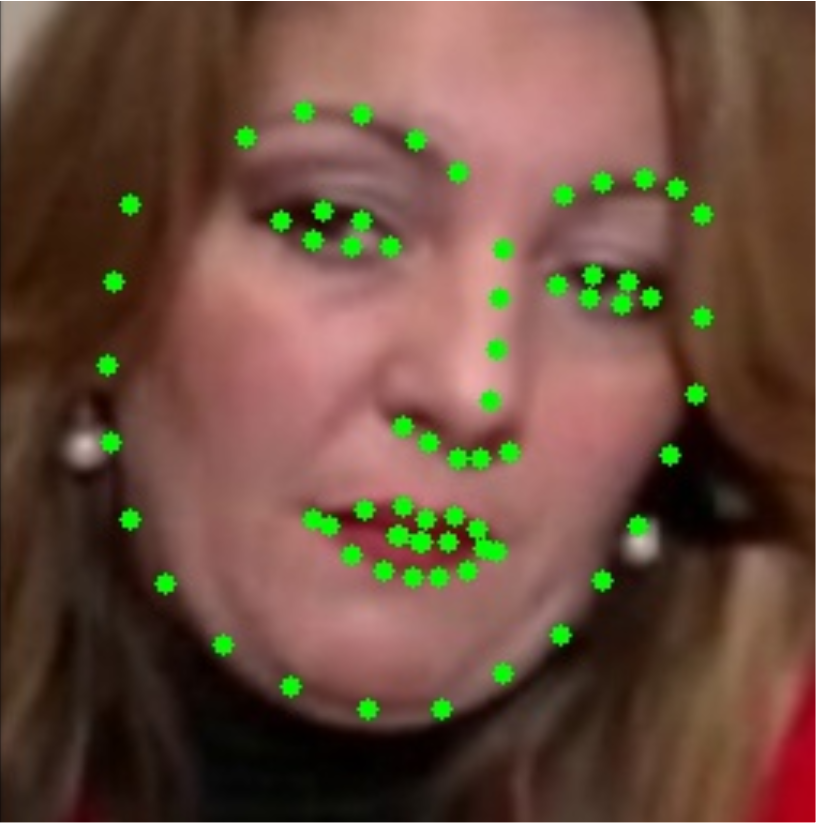}
	\end{minipage}
	\caption{Visualization of (a) detection results on the 300W testing set and (b) tracking results on the 300VW testing set.}
	\label{vis2}
\end{figure*}

Table \ref{res_300vw}  shows that the proposed tracker achieves state-of-the-art tracking accuracy on the video datasets. Among the compared methods, GAN, a combination of coordinate regression and adversarial learning, achieves the best performance after our method. Our method outperforms the NRMSE of GAN by $12.57\%$, $13.62\%$ and $7.00\%$ on the 300VW scenarios 1, 2, 3 and by $2.96\%$ on the TF dataset, respectively. 
For fine-grained analysis, we gather all testing samples in the three scenarios of 300VW and group their landmarks into five facial areas, i.e., eyes, contour, nose, eyebrows and mouth. We display the NRMSE performance on each area in Fig. \ref{dis1}. We also calculate the variances of $x$ and $y$ coordinates for each landmark and depict their average value ($s$) within each area in Fig. \ref{dis2}. From Fig. \ref{dis1}, our method decreases NRMSE on every facial area compared to GAN, especially in contour by $13.67\%$. From Fig. \ref{dis2}, we find that the contour area has the highset coordinate variances on both the $x$ and $y$ axes. 
Although GAN uses adversarial learning for spatio-temporal modeling and promotes the robustness of landmark localization in the challenging facial area, it is still sub-optimal on handling coordinates with large variances due to the intrinsic weakness of the coordinate regression technique. We significantly alleviate this problem with the proposed heatmap regression method which is more adept at capturing spatial and temporal distributions.

\section{Visualization of the Detection and Tracking Results}
We visualize the detection and tracking results of our method in Fig.  \ref{vis2}, respectively. Testing images  are sampled from the 300W and the 300VW testing sets. From these testing samples, we find that the proposed method can perform well under challenging conditions, such as occlusions, low resolution, non-frontal head poses, dramatic facial expressions and uneven illuminations.

\section{Conclusion}
To address the issues caused by the huge spatial complexity of 2D heatmaps, we propose a new method that predicts 1D heatmaps as marginal distributions on each axis to detect facial landmarks. Instead of modeling joint distribution explicitly that occupies much memory, we design a co-attention mechanism to share features between the $x$ and $y$ axes and implicitly capture joint distributions.  The light-weight 1D heatmap structure enables us to boost the output resolution to a large value despite limited GPU space, making more accurate predictions. Based on such an idea, we propose a novel landmark detector and a tracker. The detector captures spatial patterns in an image, while the tracker further captures temporal patterns in a video by a temporal refinement mechanism. Experiments on the 300W, AFLW, 300VW and TF datasets demonstrate that the proposed detector and tracker outperform state-of-the-art methods.

%
\bibliographystyle{ACM-Reference-Format}
\bibliography{arxiv}


\begin{thebibliography}{40}


\ifx \showCODEN    \undefined \def \showCODEN     #1{\unskip}     \fi
\ifx \showDOI      \undefined \def \showDOI       #1{#1}\fi
\ifx \showISBNx    \undefined \def \showISBNx     #1{\unskip}     \fi
\ifx \showISBNxiii \undefined \def \showISBNxiii  #1{\unskip}     \fi
\ifx \showISSN     \undefined \def \showISSN      #1{\unskip}     \fi
\ifx \showLCCN     \undefined \def \showLCCN      #1{\unskip}     \fi
\ifx \shownote     \undefined \def \shownote      #1{#1}          \fi
\ifx \showarticletitle \undefined \def \showarticletitle #1{#1}   \fi
\ifx \showURL      \undefined \def \showURL       {\relax}        \fi
\providecommand\bibfield[2]{#2}
\providecommand\bibinfo[2]{#2}
\providecommand\natexlab[1]{#1}
\providecommand\showeprint[2][]{arXiv:#2}

\bibitem[\protect\citeauthoryear{Asthana, Zafeiriou, Cheng, and Pantic}{Asthana
  et~al\mbox{.}}{2014}]%
        {IFA}
\bibfield{author}{\bibinfo{person}{Akshay Asthana}, \bibinfo{person}{Stefanos
  Zafeiriou}, \bibinfo{person}{Shiyang Cheng}, {and} \bibinfo{person}{Maja
  Pantic}.} \bibinfo{year}{2014}\natexlab{}.
\newblock \showarticletitle{Incremental Face Alignment in the Wild}. In
  \bibinfo{booktitle}{\emph{CVPR}}. \bibinfo{pages}{1859--1866}.
\newblock


\bibitem[\protect\citeauthoryear{Bulat and Tzimiropoulos}{Bulat and
  Tzimiropoulos}{2017}]%
        {how_far}
\bibfield{author}{\bibinfo{person}{Adrian Bulat} {and}
  \bibinfo{person}{Georgios Tzimiropoulos}.} \bibinfo{year}{2017}\natexlab{}.
\newblock \showarticletitle{How far are we from solving the 2d \& 3d face
  alignment problem?(and a dataset of 230,000 3d facial landmarks)}. In
  \bibinfo{booktitle}{\emph{ICCV}}. \bibinfo{pages}{1021--1030}.
\newblock


\bibitem[\protect\citeauthoryear{Burgos{-}Artizzu, Perona, and
  Doll{\'{a}}r}{Burgos{-}Artizzu et~al\mbox{.}}{2013}]%
        {Xavier}
\bibfield{author}{\bibinfo{person}{Xavier~P. Burgos{-}Artizzu},
  \bibinfo{person}{Pietro Perona}, {and} \bibinfo{person}{Piotr Doll{\'{a}}r}.}
  \bibinfo{year}{2013}\natexlab{}.
\newblock \showarticletitle{Robust Face Landmark Estimation under Occlusion}.
  In \bibinfo{booktitle}{\emph{ICCV}}. \bibinfo{pages}{1513--1520}.
\newblock


\bibitem[\protect\citeauthoryear{Cao, Wei, Wen, and Sun}{Cao
  et~al\mbox{.}}{2014}]%
        {ESR}
\bibfield{author}{\bibinfo{person}{Xudong Cao}, \bibinfo{person}{Yichen Wei},
  \bibinfo{person}{Fang Wen}, {and} \bibinfo{person}{Jian Sun}.}
  \bibinfo{year}{2014}\natexlab{}.
\newblock \showarticletitle{Face alignment by explicit shape regression}.
\newblock \bibinfo{journal}{\emph{IJCV}} \bibinfo{volume}{107},
  \bibinfo{number}{2} (\bibinfo{year}{2014}), \bibinfo{pages}{177--190}.
\newblock


\bibitem[\protect\citeauthoryear{Chen, Su, and Ji}{Chen et~al\mbox{.}}{2019b}]%
        {CNNCRF}
\bibfield{author}{\bibinfo{person}{Lisha Chen}, \bibinfo{person}{Hui Su}, {and}
  \bibinfo{person}{Qiang Ji}.} \bibinfo{year}{2019}\natexlab{b}.
\newblock \showarticletitle{Deep Structured Prediction for Facial Landmark
  Detection}. In \bibinfo{booktitle}{\emph{NeurIPS}}.
\newblock


\bibitem[\protect\citeauthoryear{Chen, Shen, Chen, Wei, Liu, and Yang}{Chen
  et~al\mbox{.}}{2019a}]%
        {chen2019adversarial}
\bibfield{author}{\bibinfo{person}{Y Chen}, \bibinfo{person}{C Shen},
  \bibinfo{person}{H Chen}, \bibinfo{person}{XS Wei}, \bibinfo{person}{L Liu},
  {and} \bibinfo{person}{J Yang}.} \bibinfo{year}{2019}\natexlab{a}.
\newblock \showarticletitle{Adversarial Learning of Structure-Aware Fully
  Convolutional Networks for Landmark Localization.}
\newblock \bibinfo{journal}{\emph{TPAMI}} (\bibinfo{year}{2019}).
\newblock


\bibitem[\protect\citeauthoryear{Chrysos, Antonakos, Snape, Asthana, and
  Zafeiriou}{Chrysos et~al\mbox{.}}{2018}]%
        {survey2}
\bibfield{author}{\bibinfo{person}{Grigorios~G Chrysos},
  \bibinfo{person}{Epameinondas Antonakos}, \bibinfo{person}{Patrick Snape},
  \bibinfo{person}{Akshay Asthana}, {and} \bibinfo{person}{Stefanos
  Zafeiriou}.} \bibinfo{year}{2018}\natexlab{}.
\newblock \showarticletitle{A comprehensive performance evaluation of
  deformable face tracking “In-the-Wild”}.
\newblock \bibinfo{journal}{\emph{IJCV}} \bibinfo{volume}{126},
  \bibinfo{number}{2-4} (\bibinfo{year}{2018}), \bibinfo{pages}{198--232}.
\newblock


\bibitem[\protect\citeauthoryear{Chu, Yang, Ouyang, Ma, Yuille, and Wang}{Chu
  et~al\mbox{.}}{2017}]%
        {Chu}
\bibfield{author}{\bibinfo{person}{Xiao Chu}, \bibinfo{person}{Wei Yang},
  \bibinfo{person}{Wanli Ouyang}, \bibinfo{person}{Cheng Ma},
  \bibinfo{person}{Alan~L. Yuille}, {and} \bibinfo{person}{Xiaogang Wang}.}
  \bibinfo{year}{2017}\natexlab{}.
\newblock \showarticletitle{Multi-context Attention for Human Pose Estimation}.
  In \bibinfo{booktitle}{\emph{CVPR}}. \bibinfo{pages}{5669--5678}.
\newblock


\bibitem[\protect\citeauthoryear{Dong, Yan, Ouyang, and Yang}{Dong
  et~al\mbox{.}}{2018a}]%
        {SAN}
\bibfield{author}{\bibinfo{person}{Xuanyi Dong}, \bibinfo{person}{Yan Yan},
  \bibinfo{person}{Wanli Ouyang}, {and} \bibinfo{person}{Yi Yang}.}
  \bibinfo{year}{2018}\natexlab{a}.
\newblock \showarticletitle{Style Aggregated Network for Facial Landmark
  Detection}. In \bibinfo{booktitle}{\emph{CVPR}}. \bibinfo{pages}{379--388}.
\newblock


\bibitem[\protect\citeauthoryear{Dong and Yang}{Dong and Yang}{2019}]%
        {Dong_2019_ICCV}
\bibfield{author}{\bibinfo{person}{Xuanyi Dong} {and} \bibinfo{person}{Yi
  Yang}.} \bibinfo{year}{2019}\natexlab{}.
\newblock \showarticletitle{Teacher Supervises Students How to Learn From
  Partially Labeled Images for Facial Landmark Detection}. In
  \bibinfo{booktitle}{\emph{ICCV}}.
\newblock


\bibitem[\protect\citeauthoryear{Dong, Yu, Weng, Wei, Yang, and Sheikh}{Dong
  et~al\mbox{.}}{2018b}]%
        {SBR}
\bibfield{author}{\bibinfo{person}{Xuanyi Dong}, \bibinfo{person}{Shoou{-}I
  Yu}, \bibinfo{person}{Xinshuo Weng}, \bibinfo{person}{Shih{-}En Wei},
  \bibinfo{person}{Yi Yang}, {and} \bibinfo{person}{Yaser Sheikh}.}
  \bibinfo{year}{2018}\natexlab{b}.
\newblock \showarticletitle{Supervision-by-Registration: An Unsupervised
  Approach to Improve the Precision of Facial Landmark Detectors}. In
  \bibinfo{booktitle}{\emph{CVPR}}. \bibinfo{pages}{360--368}.
\newblock


\bibitem[\protect\citeauthoryear{FGNET}{FGNET}{2014}]%
        {FGNET}
\bibfield{author}{\bibinfo{person}{FGNET}.} \bibinfo{year}{2014}\natexlab{}.
\newblock \bibinfo{title}{Talking Face Video}.
\newblock
  \bibinfo{howpublished}{\url{http://www-prima.inrialpes.fr/FGnet/data/01-TalkingFace/talking_face.html}}.
\newblock


\bibitem[\protect\citeauthoryear{Li, Wang, Zhao, and Ji}{Li
  et~al\mbox{.}}{2013}]%
        {DBN}
\bibfield{author}{\bibinfo{person}{Y. Li}, \bibinfo{person}{S. Wang},
  \bibinfo{person}{Y. Zhao}, {and} \bibinfo{person}{Q. Ji}.}
  \bibinfo{year}{2013}\natexlab{}.
\newblock \showarticletitle{Simultaneous Facial Feature Tracking and Facial
  Expression Recognition}.
\newblock \bibinfo{journal}{\emph{IEEE Transactions on Image Processing}}
  \bibinfo{volume}{22}, \bibinfo{number}{7} (\bibinfo{year}{2013}),
  \bibinfo{pages}{2559--2573}.
\newblock


\bibitem[\protect\citeauthoryear{Liu, Lu, Feng, and Zhou}{Liu
  et~al\mbox{.}}{2018}]%
        {TSTN}
\bibfield{author}{\bibinfo{person}{Hao Liu}, \bibinfo{person}{Jiwen Lu},
  \bibinfo{person}{Jianjiang Feng}, {and} \bibinfo{person}{Jie Zhou}.}
  \bibinfo{year}{2018}\natexlab{}.
\newblock \showarticletitle{Two-stream transformer networks for video-based
  face alignment}.
\newblock \bibinfo{journal}{\emph{TPAMI}} \bibinfo{volume}{40},
  \bibinfo{number}{11} (\bibinfo{year}{2018}), \bibinfo{pages}{2546--2554}.
\newblock


\bibitem[\protect\citeauthoryear{Liu, Zhu, Hu, Guo, Tang, Lei, Robertson, and
  Wang}{Liu et~al\mbox{.}}{2019}]%
        {Liu_2019_CVPR}
\bibfield{author}{\bibinfo{person}{Zhiwei Liu}, \bibinfo{person}{Xiangyu Zhu},
  \bibinfo{person}{Guosheng Hu}, \bibinfo{person}{Haiyun Guo},
  \bibinfo{person}{Ming Tang}, \bibinfo{person}{Zhen Lei},
  \bibinfo{person}{Neil~M. Robertson}, {and} \bibinfo{person}{Jinqiao Wang}.}
  \bibinfo{year}{2019}\natexlab{}.
\newblock \showarticletitle{Semantic Alignment: Finding Semantically Consistent
  Ground-Truth for Facial Landmark Detection}. In
  \bibinfo{booktitle}{\emph{CVPR}}.
\newblock


\bibitem[\protect\citeauthoryear{Lu, Yang, Batra, and Parikh}{Lu
  et~al\mbox{.}}{2016}]%
        {co-attention}
\bibfield{author}{\bibinfo{person}{Jiasen Lu}, \bibinfo{person}{Jianwei Yang},
  \bibinfo{person}{Dhruv Batra}, {and} \bibinfo{person}{Devi Parikh}.}
  \bibinfo{year}{2016}\natexlab{}.
\newblock \showarticletitle{Hierarchical Question-Image Co-Attention for Visual
  Question Answering}. In \bibinfo{booktitle}{\emph{NIPS}}.
  \bibinfo{pages}{289--297}.
\newblock


\bibitem[\protect\citeauthoryear{Miao, Zhen, Liu, Deng, Athitsos, and
  Huang}{Miao et~al\mbox{.}}{2018}]%
        {DSRN}
\bibfield{author}{\bibinfo{person}{Xin Miao}, \bibinfo{person}{Xiantong Zhen},
  \bibinfo{person}{Xianglong Liu}, \bibinfo{person}{Cheng Deng},
  \bibinfo{person}{Vassilis Athitsos}, {and} \bibinfo{person}{Heng Huang}.}
  \bibinfo{year}{2018}\natexlab{}.
\newblock \showarticletitle{Direct Shape Regression Networks for End-to-End
  Face Alignment}. In \bibinfo{booktitle}{\emph{CVPR}}.
  \bibinfo{pages}{5040--5049}.
\newblock


\bibitem[\protect\citeauthoryear{Newell, Yang, and Deng}{Newell
  et~al\mbox{.}}{2016}]%
        {hourglass}
\bibfield{author}{\bibinfo{person}{Alejandro Newell}, \bibinfo{person}{Kaiyu
  Yang}, {and} \bibinfo{person}{Jia Deng}.} \bibinfo{year}{2016}\natexlab{}.
\newblock \showarticletitle{Stacked Hourglass Networks for Human Pose
  Estimation}. In \bibinfo{booktitle}{\emph{ECCV}}. \bibinfo{pages}{483--499}.
\newblock


\bibitem[\protect\citeauthoryear{Nibali, He, Morgan, and Prendergast}{Nibali
  et~al\mbox{.}}{2018}]%
        {better_heatmap2}
\bibfield{author}{\bibinfo{person}{Aiden Nibali}, \bibinfo{person}{Zhen He},
  \bibinfo{person}{Stuart Morgan}, {and} \bibinfo{person}{Luke Prendergast}.}
  \bibinfo{year}{2018}\natexlab{}.
\newblock \showarticletitle{Numerical Coordinate Regression with Convolutional
  Neural Networks}.
\newblock \bibinfo{journal}{\emph{CoRR}}  \bibinfo{volume}{abs/1801.07372}
  (\bibinfo{year}{2018}).
\newblock


\bibitem[\protect\citeauthoryear{Peng, Feris, Wang, and Metaxas}{Peng
  et~al\mbox{.}}{2016}]%
        {encoder_decoder}
\bibfield{author}{\bibinfo{person}{Xi Peng}, \bibinfo{person}{Rogerio~S Feris},
  \bibinfo{person}{Xiaoyu Wang}, {and} \bibinfo{person}{Dimitris~N Metaxas}.}
  \bibinfo{year}{2016}\natexlab{}.
\newblock \showarticletitle{A recurrent encoder-decoder network for sequential
  face alignment}. In \bibinfo{booktitle}{\emph{ECCV}}.
  \bibinfo{pages}{38--56}.
\newblock


\bibitem[\protect\citeauthoryear{Qian, Sun, Wu, Qian, and Jia}{Qian
  et~al\mbox{.}}{2019}]%
        {Qian_2019_ICCV}
\bibfield{author}{\bibinfo{person}{Shengju Qian}, \bibinfo{person}{Keqiang
  Sun}, \bibinfo{person}{Wayne Wu}, \bibinfo{person}{Chen Qian}, {and}
  \bibinfo{person}{Jiaya Jia}.} \bibinfo{year}{2019}\natexlab{}.
\newblock \showarticletitle{Aggregation via Separation: Boosting Facial
  Landmark Detector With Semi-Supervised Style Translation}. In
  \bibinfo{booktitle}{\emph{ICCV}}.
\newblock


\bibitem[\protect\citeauthoryear{Ren, Cao, Wei, and Sun}{Ren
  et~al\mbox{.}}{2014}]%
        {Ren}
\bibfield{author}{\bibinfo{person}{Shaoqing Ren}, \bibinfo{person}{Xudong Cao},
  \bibinfo{person}{Yichen Wei}, {and} \bibinfo{person}{Jian Sun}.}
  \bibinfo{year}{2014}\natexlab{}.
\newblock \showarticletitle{Face Alignment at 3000 {FPS} via Regressing Local
  Binary Features}. In \bibinfo{booktitle}{\emph{CVPR}}.
  \bibinfo{pages}{1685--1692}.
\newblock


\bibitem[\protect\citeauthoryear{Robinson, Li, Zhang, Fu, and
  Tulyakov}{Robinson et~al\mbox{.}}{2019}]%
        {Robinson_2019_ICCV}
\bibfield{author}{\bibinfo{person}{Joseph~P. Robinson},
  \bibinfo{person}{Yuncheng Li}, \bibinfo{person}{Ning Zhang},
  \bibinfo{person}{Yun Fu}, {and} \bibinfo{person}{Sergey Tulyakov}.}
  \bibinfo{year}{2019}\natexlab{}.
\newblock \showarticletitle{Laplace Landmark Localization}. In
  \bibinfo{booktitle}{\emph{ICCV}}.
\newblock


\bibitem[\protect\citeauthoryear{Sagonas, Antonakos, Tzimiropoulos, Zafeiriou,
  and Pantic}{Sagonas et~al\mbox{.}}{2016}]%
        {sagonas2016300}
\bibfield{author}{\bibinfo{person}{Christos Sagonas},
  \bibinfo{person}{Epameinondas Antonakos}, \bibinfo{person}{Georgios
  Tzimiropoulos}, \bibinfo{person}{Stefanos Zafeiriou}, {and}
  \bibinfo{person}{Maja Pantic}.} \bibinfo{year}{2016}\natexlab{}.
\newblock \showarticletitle{300 faces in-the-wild challenge: Database and
  results}.
\newblock \bibinfo{journal}{\emph{Image and vision computing}}
  \bibinfo{volume}{47} (\bibinfo{year}{2016}), \bibinfo{pages}{3--18}.
\newblock


\bibitem[\protect\citeauthoryear{Shen, Zafeiriou, Chrysos, Kossaifi,
  Tzimiropoulos, and Pantic}{Shen et~al\mbox{.}}{2015}]%
        {shen2015first}
\bibfield{author}{\bibinfo{person}{Jie Shen}, \bibinfo{person}{Stefanos
  Zafeiriou}, \bibinfo{person}{Grigoris~G Chrysos}, \bibinfo{person}{Jean
  Kossaifi}, \bibinfo{person}{Georgios Tzimiropoulos}, {and}
  \bibinfo{person}{Maja Pantic}.} \bibinfo{year}{2015}\natexlab{}.
\newblock \showarticletitle{The first facial landmark tracking in-the-wild
  challenge: Benchmark and results}. In \bibinfo{booktitle}{\emph{ICCV
  Workshops}}. \bibinfo{pages}{50--58}.
\newblock


\bibitem[\protect\citeauthoryear{Simonyan and Zisserman}{Simonyan and
  Zisserman}{2014}]%
        {TSCN}
\bibfield{author}{\bibinfo{person}{Karen Simonyan} {and}
  \bibinfo{person}{Andrew Zisserman}.} \bibinfo{year}{2014}\natexlab{}.
\newblock \showarticletitle{Two-stream convolutional networks for action
  recognition in videos}. In \bibinfo{booktitle}{\emph{NIPS}}.
  \bibinfo{pages}{568--576}.
\newblock


\bibitem[\protect\citeauthoryear{Sun, Wu, Liu, Yang, Wang, Zhou, Ye, and
  Qian}{Sun et~al\mbox{.}}{2019}]%
        {Sun_2019_ICCV}
\bibfield{author}{\bibinfo{person}{Keqiang Sun}, \bibinfo{person}{Wayne Wu},
  \bibinfo{person}{Tinghao Liu}, \bibinfo{person}{Shuo Yang},
  \bibinfo{person}{Quan Wang}, \bibinfo{person}{Qiang Zhou},
  \bibinfo{person}{Zuochang Ye}, {and} \bibinfo{person}{Chen Qian}.}
  \bibinfo{year}{2019}\natexlab{}.
\newblock \showarticletitle{FAB: A Robust Facial Landmark Detection Framework
  for Motion-Blurred Videos}. In \bibinfo{booktitle}{\emph{ICCV}}.
\newblock


\bibitem[\protect\citeauthoryear{Sun, Xiao, Wei, Liang, and Wei}{Sun
  et~al\mbox{.}}{2018}]%
        {better_heatmap1}
\bibfield{author}{\bibinfo{person}{Xiao Sun}, \bibinfo{person}{Bin Xiao},
  \bibinfo{person}{Fangyin Wei}, \bibinfo{person}{Shuang Liang}, {and}
  \bibinfo{person}{Yichen Wei}.} \bibinfo{year}{2018}\natexlab{}.
\newblock \showarticletitle{Integral Human Pose Regression}. In
  \bibinfo{booktitle}{\emph{ECCV}}. \bibinfo{pages}{536--553}.
\newblock


\bibitem[\protect\citeauthoryear{Sun, Wang, and Tang}{Sun
  et~al\mbox{.}}{2013}]%
        {Cascade_CNN}
\bibfield{author}{\bibinfo{person}{Yi Sun}, \bibinfo{person}{Xiaogang Wang},
  {and} \bibinfo{person}{Xiaoou Tang}.} \bibinfo{year}{2013}\natexlab{}.
\newblock \showarticletitle{Deep Convolutional Network Cascade for Facial Point
  Detection}. In \bibinfo{booktitle}{\emph{CVPR}}. \bibinfo{pages}{3476--3483}.
\newblock


\bibitem[\protect\citeauthoryear{Tai, Liang, Liu, Duan, Li, Wang, Huang, and
  Chen}{Tai et~al\mbox{.}}{2019}]%
        {tai2018towards}
\bibfield{author}{\bibinfo{person}{Ying Tai}, \bibinfo{person}{Yicong Liang},
  \bibinfo{person}{Xiaoming Liu}, \bibinfo{person}{Lei Duan},
  \bibinfo{person}{Jilin Li}, \bibinfo{person}{Chengjie Wang},
  \bibinfo{person}{Feiyue Huang}, {and} \bibinfo{person}{Yu Chen}.}
  \bibinfo{year}{2019}\natexlab{}.
\newblock \showarticletitle{Towards highly accurate and stable face alignment
  for high-resolution videos}. In \bibinfo{booktitle}{\emph{AAAI}}.
  \bibinfo{pages}{8893--8900}.
\newblock


\bibitem[\protect\citeauthoryear{Vaswani, Shazeer, Parmar, Uszkoreit, Jones,
  Gomez, Kaiser, and Polosukhin}{Vaswani et~al\mbox{.}}{2017}]%
        {transformer}
\bibfield{author}{\bibinfo{person}{Ashish Vaswani}, \bibinfo{person}{Noam
  Shazeer}, \bibinfo{person}{Niki Parmar}, \bibinfo{person}{Jakob Uszkoreit},
  \bibinfo{person}{Llion Jones}, \bibinfo{person}{Aidan~N. Gomez},
  \bibinfo{person}{Lukasz Kaiser}, {and} \bibinfo{person}{Illia Polosukhin}.}
  \bibinfo{year}{2017}\natexlab{}.
\newblock \showarticletitle{Attention is All you Need}. In
  \bibinfo{booktitle}{\emph{NIPS}}. \bibinfo{pages}{5998--6008}.
\newblock


\bibitem[\protect\citeauthoryear{Wu, Qian, Yang, Wang, Cai, and Zhou}{Wu
  et~al\mbox{.}}{2018}]%
        {LAB}
\bibfield{author}{\bibinfo{person}{Wayne Wu}, \bibinfo{person}{Chen Qian},
  \bibinfo{person}{Shuo Yang}, \bibinfo{person}{Quan Wang},
  \bibinfo{person}{Yici Cai}, {and} \bibinfo{person}{Qiang Zhou}.}
  \bibinfo{year}{2018}\natexlab{}.
\newblock \showarticletitle{Look at Boundary: {A} Boundary-Aware Face Alignment
  Algorithm}. In \bibinfo{booktitle}{\emph{CVPR}}. \bibinfo{pages}{2129--2138}.
\newblock


\bibitem[\protect\citeauthoryear{Wu and Ji}{Wu and Ji}{2019}]%
        {survey}
\bibfield{author}{\bibinfo{person}{Yue Wu} {and} \bibinfo{person}{Qiang Ji}.}
  \bibinfo{year}{2019}\natexlab{}.
\newblock \showarticletitle{Facial Landmark Detection: {A} Literature Survey}.
\newblock \bibinfo{journal}{\emph{IJCV}} \bibinfo{volume}{127},
  \bibinfo{number}{2} (\bibinfo{year}{2019}), \bibinfo{pages}{115--142}.
\newblock


\bibitem[\protect\citeauthoryear{Wu, Wang, and Ji}{Wu et~al\mbox{.}}{2014}]%
        {RBM}
\bibfield{author}{\bibinfo{person}{Yue Wu}, \bibinfo{person}{Ziheng Wang},
  {and} \bibinfo{person}{Qiang Ji}.} \bibinfo{year}{2014}\natexlab{}.
\newblock \showarticletitle{A Hierarchical Probabilistic Model for Facial
  Feature Detection}. In \bibinfo{booktitle}{\emph{CVPR}}.
  \bibinfo{pages}{1781--1788}.
\newblock


\bibitem[\protect\citeauthoryear{Xiong and De~la Torre}{Xiong and De~la
  Torre}{2013}]%
        {SDM}
\bibfield{author}{\bibinfo{person}{Xuehan Xiong} {and}
  \bibinfo{person}{Fernando De~la Torre}.} \bibinfo{year}{2013}\natexlab{}.
\newblock \showarticletitle{Supervised descent method and its applications to
  face alignment}. In \bibinfo{booktitle}{\emph{CVPR}}.
  \bibinfo{pages}{532--539}.
\newblock


\bibitem[\protect\citeauthoryear{Yin, Wang, Peng, Chen, and Pan}{Yin
  et~al\mbox{.}}{2019}]%
        {GAN_Tracking}
\bibfield{author}{\bibinfo{person}{Shi Yin}, \bibinfo{person}{Shangfei Wang},
  \bibinfo{person}{Guozhu Peng}, \bibinfo{person}{Xiaoping Chen}, {and}
  \bibinfo{person}{Bowen Pan}.} \bibinfo{year}{2019}\natexlab{}.
\newblock \showarticletitle{Capturing Spatial and Temporal Patterns for Facial
  Landmark Tracking through Adversarial Learning}. In
  \bibinfo{booktitle}{\emph{IJCAI}}. \bibinfo{pages}{1010--1017}.
\newblock


\bibitem[\protect\citeauthoryear{Zhang, Zhu, Dai, Ye, and Zhu}{Zhang
  et~al\mbox{.}}{2020}]%
        {better_heatmap3}
\bibfield{author}{\bibinfo{person}{Feng Zhang}, \bibinfo{person}{Xiatian Zhu},
  \bibinfo{person}{Hanbin Dai}, \bibinfo{person}{Mao Ye}, {and}
  \bibinfo{person}{Ce Zhu}.} \bibinfo{year}{2020}\natexlab{}.
\newblock \showarticletitle{Distribution-Aware Coordinate Representation for
  Human Pose Estimation}.
\newblock \bibinfo{journal}{\emph{CVPR}} (\bibinfo{year}{2020}).
\newblock


\bibitem[\protect\citeauthoryear{Zhang, Luo, Loy, and Tang}{Zhang
  et~al\mbox{.}}{2016}]%
        {TCDCN}
\bibfield{author}{\bibinfo{person}{Zhanpeng Zhang}, \bibinfo{person}{Ping Luo},
  \bibinfo{person}{Chen~Change Loy}, {and} \bibinfo{person}{Xiaoou Tang}.}
  \bibinfo{year}{2016}\natexlab{}.
\newblock \showarticletitle{Learning deep representation for face alignment
  with auxiliary attributes}.
\newblock \bibinfo{journal}{\emph{TPAMI}} \bibinfo{volume}{38},
  \bibinfo{number}{5} (\bibinfo{year}{2016}), \bibinfo{pages}{918--930}.
\newblock


\bibitem[\protect\citeauthoryear{Zhu, Shi, Zheng, and Sadiq}{Zhu
  et~al\mbox{.}}{2019}]%
        {Zhu_2019_CVPR}
\bibfield{author}{\bibinfo{person}{Meilu Zhu}, \bibinfo{person}{Daming Shi},
  \bibinfo{person}{Mingjie Zheng}, {and} \bibinfo{person}{Muhammad Sadiq}.}
  \bibinfo{year}{2019}\natexlab{}.
\newblock \showarticletitle{Robust Facial Landmark Detection via
  Occlusion-Adaptive Deep Networks}. In \bibinfo{booktitle}{\emph{CVPR}}.
\newblock


\bibitem[\protect\citeauthoryear{Zhu, Li, Change~Loy, and Tang}{Zhu
  et~al\mbox{.}}{2015}]%
        {CFSS}
\bibfield{author}{\bibinfo{person}{Shizhan Zhu}, \bibinfo{person}{Cheng Li},
  \bibinfo{person}{Chen Change~Loy}, {and} \bibinfo{person}{Xiaoou Tang}.}
  \bibinfo{year}{2015}\natexlab{}.
\newblock \showarticletitle{Face alignment by coarse-to-fine shape searching}.
  In \bibinfo{booktitle}{\emph{CVPR}}. \bibinfo{pages}{4998--5006}.
\newblock


\end{thebibliography}

%

\end{document}